%% file: main.tex
\documentclass[sigconf, screen]{acmart}

\AtBeginDocument{%
  }

\setcopyright{acmlicensed}
\acmPrice{15.00}
\acmDOI{10.1145/3533767.3534223}
\acmYear{2022}
\copyrightyear{2022}
\acmSubmissionID{issta22main-p337-p}
\acmISBN{978-1-4503-9379-9/22/07}
\acmConference[ISSTA '22]{Proceedings of the 31st ACM SIGSOFT International Symposium on Software Testing and Analysis}{July 18--22, 2022}{Virtual, South Korea}
\acmBooktitle{Proceedings of the 31st ACM SIGSOFT International Symposium on Software Testing and Analysis (ISSTA '22), July 18--22, 2022, Virtual, South Korea}

\input{macro}

\begin{document}

\title{Detecting Multi-sensor Fusion Errors in Advanced Driver-Assistance Systems}
\author{Ziyuan Zhong}
\email{ziyuan.zhong@columbia.edu}
\affiliation{%
  \institution{Columbia University}
  \country{United States}
}

\author{Zhisheng Hu}
\email{zhishenghu@baidu.com}
\affiliation{%
  \institution{Baidu Security}
  \country{United States}
}

\author{Shengjian Guo}
\email{sjguo@baidu.com}
\affiliation{%
  \institution{Baidu Security}
  \country{United States}
}

\author{Xinyang Zhang}
\email{xinyangzhang@baidu.com}
\affiliation{%
  \institution{Baidu Security}
  \country{United States}
}

\author{Zhenyu Zhong}
\email{edwardzhong@baidu.com}
\affiliation{%
  \institution{Baidu Security}
  \country{United States}
}

\author{Baishakhi Ray}
\email{rayb@cs.columbia.edu}
\affiliation{%
  \institution{Columbia University}
  \country{United States}
}

\renewcommand{\shortauthors}{Ziyuan Zhong, Zhisheng Hu, Shengjian Guo, Xinyang Zhang, Zhenyu Zhong, and Baishakhi Ray}

\input{body/abstract.tex}




\maketitle

\input{body/1_intro}
\input{body/2_fusion_background}
\input{body/3_overview}
\input{body/4_fuzzing}
\input{evaluation/rq1}

\input{evaluation/rq2}

\input{evaluation/rq3}
\input{body/8_discussion}
\input{body/9_conclusion}
\input{body/10_acknowledgement}
\balance

\bibliographystyle{ACM-Reference-Format}
\bibliography{main}
\clearpage
\appendix
\input{body/appendices}

\end{document}

%% file: macro.tex
\usepackage{color}
\usepackage{microtype}

\newcommand{\newedit}[1]{\textcolor{black}{#1}}
\newcommand{\newnewedit}[1]{\textcolor{black}{#1}}
\newcommand{\new}[1]{\textcolor{black} {#1}}

\newcommand{\credit}[1]{\textcolor{black}{#1}}
\newcommand{\extend}[1]{}
\usepackage{algorithm}
\usepackage{algorithmic}

\usepackage{newfloat}
\usepackage{listings}
\floatstyle{ruled}
\newfloat{listing}{tb}{lst}{}
\floatname{listing}{Listing}

\usepackage{subcaption}
\usepackage{xspace}
\usepackage{booktabs}
\usepackage{amsmath}
\usepackage{amsthm}
\usepackage{lipsum}
\usepackage{cleveref}
\theoremstyle{definition}
\newtheorem{defi}{Definition}
\usepackage{framed}
\usepackage{enumitem}
\usepackage{dsfont}
\usepackage[normalem]{ulem}
\usepackage{balance}

\newcommand{\tool}{\textit{FusED}\xspace}
\newcommand{\carla}{\textsc{carla}\xspace}
\newcommand{\CARLA}{\textsc{carla}\xspace}
\newcommand{\tm}{\textsc{carla traffic manager}\xspace}
\newcommand{\op}{\textsc{Openpilot}\xspace}

\newcommand{\ad}{\hbox{\textsc{AD}}\xspace}
\newcommand{\ads}{\hbox{\textsc{ADS}}\xspace}
\newcommand{\adas}{\hbox{\textsc{ADAS}}\xspace}
\newcommand{\msf}{\hbox{\textsc{MSF}}\xspace}

\newcommand{\original}{\textsc{default}\xspace}

\newcommand{\mathwork}{\textsc{mathworks}\xspace}
\newcommand{\mathworkoriginal}{\textsc{mathworks}\xspace}

\newcommand{\mathworkmoving}{\textsc{mathworks+}\xspace}

\newcommand{\nfbugs}{non-fusion errors\xspace}

\newcommand{\bugs}{fusion errors\xspace}
\newcommand{\bug}{fusion error\xspace}
\newcommand{\failure}{failure\xspace}
\newcommand{\failures}{failures\xspace}
\newcommand{\fusionfaults}{fusion faults\xspace}
\newcommand{\fusionfault}{fusion fault\xspace}

\newcommand{\RA}{\textsc{random}\xspace}
\newcommand{\GA}{\textsc{ga}\xspace}
\newcommand{\alg}{\textsc{ga-fusion}\xspace}

\newcommand{\Sa}{{S1}\xspace}
\newcommand{\Sb}{{S2}\xspace}

\newcommand{\lss}{{driving environments}\xspace}

\newcommand{\specificss}{{scenarios}\xspace}

\newcommand{\lead}{{lead}\xspace}
\newcommand{\leads}{{leads}\xspace}

\newcommand{\RS}[2]{%
\begin{framed}%
\vspace{-1mm}
\noindent
\textbf{Result {#1}:~}{\emph {#2}}%
\vspace{-1mm}
\end{framed}
}

\renewcommand{\sout}[1]{}

%% file: body/abstract.tex
\begin{abstract}
Advanced Driver-Assistance Systems (\adas) have been thriving and widely deployed in recent years.
In general, these systems receive sensor data, compute driving decisions, and output control signals to the vehicles. To smooth out the uncertainties brought by sensor \newedit{outputs}\sout{inputs}, they usually leverage \textbf{multi-sensor fusion} (\msf) to fuse the sensor \newedit{outputs}\sout{inputs} and produce 
a more reliable understanding of the surroundings. 
However, \msf cannot completely eliminate the uncertainties since it lacks the knowledge about which sensor provides the most accurate data and how to optimally integrate the data provided by the sensors.
As a result, critical consequences might happen unexpectedly.
In this work, we observed that the popular \msf methods in an industry-grade \adas can mislead the car control 
and result in serious safety hazards. 
\sout{Misbehavior can happen regardless 
of the used fusion methods and the accurate data from at least one sensor.}We \sout{call such}\newedit{define the \failures (e.g., car crashes) caused by the faulty \msf}\sout{errors} as \bugs and develop a novel evolutionary-based domain-specific search framework, \tool, for the efficient detection of \bugs. We further apply causality analysis to show that the found \bugs are indeed caused by the \msf method.
We evaluate our framework on two widely used \msf methods in two driving environments.
Experimental results show that \tool identifies more than 150 \bugs. Finally, we provide several suggestions to improve the \msf methods we study.
\end{abstract}

%% file: body/1_intro.tex
\section{Introduction}

Advanced Driver-Assistance Systems (\adas) are human-machine systems that assist drivers in driving and parking functions and have been widely deployed on production passenger vehicles \cite{adas20} (e.g. 
Tesla's AutoPilot\sout{, Cadillac's Super Cruise,} and Comma Two's \op\cite{openpilotconsumerreport}). 
Unlike the full automation promised by so-called self-driving cars, \adas provides partial automation like adaptive cruise control, lane departure warning\sout{, traffic signals recognition}, etc., to promote a safe and effortless driving experience. 
Although \adas are developed to increase road safety,  they can malfunction and lead to critical consequences\cite{tesladeaths}. It is thus important to improve the reliability of \adas. 

\begin{figure}[ht]
\centering
    {\includegraphics[width=0.45\textwidth]{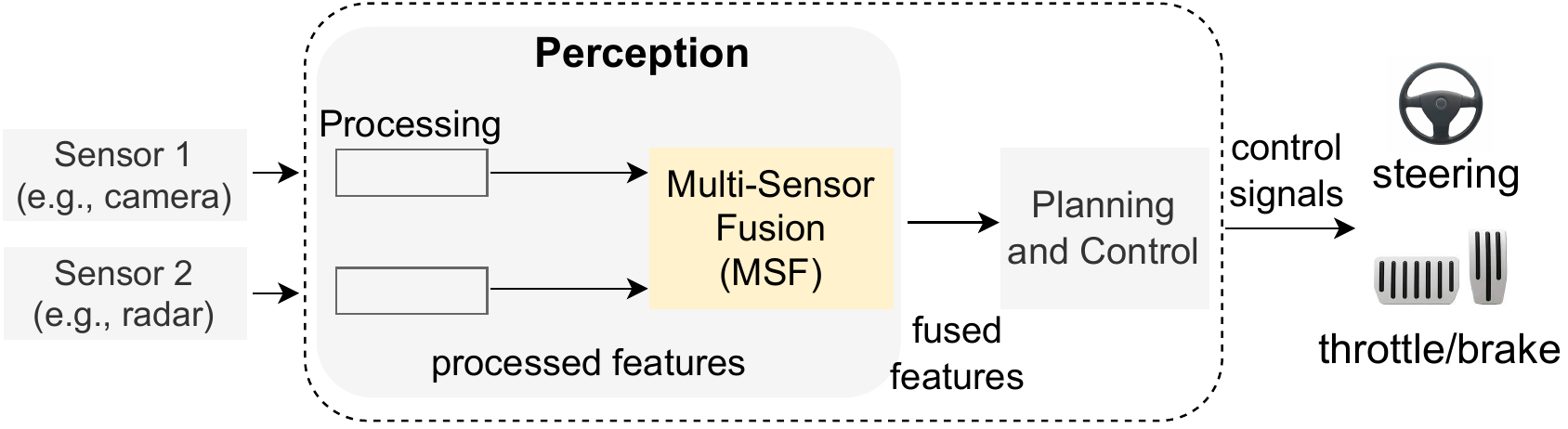}}
\caption{\small{The architecture of a typical \adas system.}}
\label{fig:fusion_highlevel}
\vspace{-8mm}
\end{figure}

A typical \adas, as shown in \Cref{fig:fusion_highlevel}, takes inputs from a set of sensors (e.g., camera, radar, etc.) and outputs driving decisions to the controlled vehicle. It usually has a perception module that interprets the sensor data to understand the surroundings, a planning module that plans the vehicle's successive trajectory, and a control module that makes concrete actuator control signals to drive the vehicle. 
Oftentimes individual sensor data could be unreliable under various extreme environments. For example, a camera can fail miserably in a dark environment, in which a radar can function correctly. 
In contrast, a radar can miss some small moving objects due to its low resolution, while a camera usually provides precise measurements in such cases.
To enable an \adas to drive reliably in most environments, researchers have adopted complementary sensors and developed multi-sensor fusion (\msf) methods to aggregate the data from multiple sensors to model the environment more reliably. If one sensor fails, \msf can still work with other sensors to provide reliable information for the downstream modules and enable the \adas to operate safely.

However, an \msf hardly knows which sensor \newedit{output}\sout{input} to rely on at each time step. Thus, it neither thoroughly eliminates the uncertainty nor always weighs more on the correct sensor data. This inherent flaw may introduce safety risks to the \adas. 
In this paper, we study a popular commercial \adas named \op and show that sometimes its \msf can mistakenly prioritize faulty sensor information over the correct ones.  
Such incorrect fusion logic can lead the vehicle to critical accidents.
To this end, this paper focuses on automatically detecting accidents (i.e., collisions) that can occur due to incorrect fusion logic---we call such accidents as {\em fusion errors}. 

Similar to existing \adas testing~\cite{interactiontest, testing_vision18, 10.1145/3395363.3397386}, we resort to simulation rather than real-world testing as the latter is prohibitively expensive. A vehicle controlled by the \adas{}\newedit{, a.k.a.}\sout{, also known as} the \emph{ego car}, drives through the scenario generated by a\sout{ high-fidelity} simulator. Here, the MSF logic of the \adas is under test. The semantic validity of the generated scenarios are guaranteed by the usage of the simulator's traffic manager which controls other vehicles (not controlled by \adas) to behave in a realistic way.
Our aim is to simulate scenarios that facilitate fusion errors\sout{ in the \adas} as  detecting fusion errors even in a simulated environment is challenging.




\noindent
\textbf{Challenges.} 
There are two main challenges in simulating and detecting \bugs:  
\begin{enumerate}[leftmargin=*,label={\Roman*.}]
    \item 
    \newedit{The \failure (i.e., collision) cases in a commercial-grade \adas are rare since it functions properly most of time. 
The \failure cases caused by the fusion method are even more sparse since there can be other causes of failure like the malfunctions of all the sensors}\sout{ existing fusion methods can function properly most of the time, so their failure cases are sparse among all the driving situations}.
Given testing an \newedit{\adas}\sout{\ad system} is costly, it is non-trivial to identify \newedit{these fusion-induced}\sout{sparsed} failure cases within a limited time budget. 
    \item
    Even if we detect \newedit{a failure}\sout{an accident}, it is hard to conclude its root cause is an incorrect fusion logic. Employing simple differential testing (i.e., we simulate the whole driving scenario with alternative fusion logic and avoid the collision) cannot say with certainty that the root cause was the faulty fusion logic. This is because many uncertainties are involved in the \newedit{simulation process}\sout{end-to-end \adas system}---non-deterministic sensor \newedit{outputs}\sout{inputs}, random time delays between simulator and \newedit{\adas}\sout{controller}, etc.; reproducing the exact collision is non-trivial. 
\end{enumerate}



\noindent
\textbf{Our Approach.} 
\newedit{
We treat a \failure (i.e., collision) caused by the faulty fusion method as a \bug.
A reasonable assumption is that a \emph{fusion fault} occurs as the fusion method chooses 
 a wrong sensor output while a correct output from another sensor was available. 
Consequently, a \emph{\bug} usually takes place when (i) some \emph{\fusionfaults} happen, and (ii) a \emph{failure} (i.e., ego car collision) takes place. To detect \bugs, we first use fuzz testing with objectives promoting the occurrence of many \fusionfaults and the resulting failure.}
\sout{A \bug takes place (i) when the output of two sensors differs significantly, and 
(ii) the fusion logic chooses the wrong sensor output even though the correct output is available. 
To detect such errors, we first use fuzz testing that promotes two sensors to behave differently under the same environment.} 
If a \newedit{failure}\sout{crash} happens, we further apply root cause analysis to filter out the \newedit{failures}\sout{errors} that may not have been caused by faulty fusion logic.
These two steps, as detailed below, are carefully designed and implemented with a tool, \tool~(Fusion Error Detector), to address the challenges mentioned above. 

\textit{Step-I:~Fuzzing}.
To induce \bugs, the simulator needs to generate scenarios that promote the fusion component to provide inaccurate prediction although the non-chosen sensor output provides accurate prediction\newedit{, and lead the ego car to collision}.
In the driving automation testing domain, recent works leverage fuzz testing to simulate input scenarios in which an ego car runs and the fuzzer is optimized to search for failure-inducing scenarios ~\cite{testing_vision18, avfuzzer, covavfuzz, autofuzz21}. However, these methods treat the ego-car 
system 
as a black-box and ignore the attainable run-time information of the system. Inspired by the grey-box fuzzing of traditional software fuzzing literature \cite{fuzzing_survey}, we propose an evolutionary algorithm-based fuzzing that utilizes the input and output information of the fusion component of the system. 
In particular, \newedit{to promote the fusion component to make more \fusionfaults,} we propose a novel objective function that maximizes the difference between the fusion component's prediction and the ground-truth, while minimizing the difference between the most accurate sensor's prediction and the ground-truth. Here, ground-truth is the actual relative location and relative speed of the leading vehicle w.r.t. the ego-car. \newedit{To promote the ego car's crash, similar to previous works \cite{testing_vision18, avfuzzer, autofuzz21}, we use an objective minimizing the ego car's distance to its leading vehicle. The two objectives synergistically promote finding scenarios that trigger \bugs.}
\sout{If the simulation then witnesses a collision of the ego car, it is highly likely the fusion logic chooses the wrong output over the correct one.}

\textit{Step-II:~Root Cause Analysis.} To address challenge-II, i.e., to check whether the observed \newedit{failure}\sout{collision} is indeed due to the fusion logic, we study if the \newedit{failure}\sout{error} still happens after choosing an alternate fusion logic in an otherwise identical simulation environment. \newedit{Here, we intend to do a controlled study to measure the effect of faulty fusion logic.} However, maintaining an identical setting is infeasible because of many uncertainties and randomness in the \newedit{environment of} simulator and controller\sout{ environment}. Thus, we rely on the theory of causal analysis. 
Based on \sout{domain knowledge}\newedit{the understanding of the studied \adas and the simulator}, we construct a causal graph, where graph nodes are all the variables that can influence the occurrence of a collision during a simulation, and the edges are links that show their influence with each other.
We then intervene and change the fusion logic by keeping all the other nodes identical in the causal graph. 
Such intervention is applied by setting the communications between the simulator and ego-car deterministic and synchronous for all the simulations. To efficiently find a fusion method that can avoid the collision, we use a \emph{best-sensor fusion} method, which always selects the sensor's output that is closest to the ground truth.
If we no longer see the collision in this counterfactual world, we conclude that the root cause of the observed collision was incorrect fusion logic. Otherwise, we discard the \newedit{failure}\sout{error} observed during fuzzing.  
To further reduce double-counting the same \newedit{\bug}\sout{error}, we propose a new\sout{ \bugs} counting metric based on the coverage of the ego car's location and speed trajectory during each simulation.

To the best of our knowledge, our technique is the first fuzzing method targeting the \adas fusion component. \new{In total, \tool has found more than 150 \bugs.}
In summary, we make the following contributions:
\begin{itemize}[leftmargin=*]
\item We define \emph{\bugs} and develop a novel grey-box fuzzing technique for efficiently revealing the  \emph{\bugs} in \adas.
\item We analyze the\sout{ root} causes of the \emph{\bugs} using \newedit{causal}\sout{causality} analysis.
\item We evaluate \tool 
in an industry-grade \adas, and show that it can disclose safety issues. 
\item We propose suggestions to mitigate \emph{\bugs} and effectively reduce  \emph{\bugs} in a preliminary study.
\end{itemize}

The source code of our tool and interesting findings are available at \url{https://github.com/AIasd/FusED}.

%% file: body/2_fusion_background.tex
\section{Background: Fusion in \adas}
\label{sec:background}


The "Standard Road Motor Vehicle Driving Automation System Classification and Definition"~\cite{saeadlevels} 
categorizes driving automation systems into six levels. 
Advanced Driver-{\em Assistance} Systems (\adas) usually consists of levels 0 to 2, which only provides temporary intervention (e.g., Autonomous Emergency Braking (AEB)) or longitudianl/latitudinal control (e.g., Automated Lane Centering (ALC) and Adaptive Cruise Control (ACC)) while requiring the driver's attention all the time. 
In contrast, Automated Driving Systems (\ads) consist of levels 3 to 5, which allow the driver to not pay attention all the time.
In this section, we introduce commonly used fusion methods and related errors for driving automation systems.
In particular, we focus on \op, a level2 industry-grade \adas. However, we believe our approach can also generalize to \ads that use similar fusion methods~\cite{apollo, autoware}.

We next define the terminologies used later.

\begin{itemize}[leftmargin=*]
    \item[--] 
A \emph{driving environment} is a parameterized space where search during the fuzzing will be bounded.
    \item[--] 
A \emph{scenario} is a concrete instance in the driving environment. 
    \item[--] 
The \emph{ego car} is the vehicle controlled by the \adas under test. 
    \item[--] 
The \emph{NPC (non-player character) vehicles} are the vehicles other than the ego car\sout{ in the scenario}.
    \item[--] 
The \emph{leading vehicle} is the vehicle ahead of the ego car in the same lane. 
    \item[--]  
A high-fidelity \emph{simulator} provides an end-to-end simulation environment for testing \adas. It generates sensor data at regular intervals (from cameras, radar, etc.) that can be fed into the \adas under test, and receives control signal from the \adas to update the ego car in the simulated world.
\end{itemize}



\subsection{Fusion in Driving Automation}
\label{ssec:fusion_adas}

Most industry-grade driving automation systems, including \adas and \ads, leverage multi-sensor fusion (\msf) to avoid potential accidents caused by the failure of a single sensor \cite{apollo, autoware, openpilot}. \msf often works with camera and radar, camera and Lidar, or the combination of camera, radar, and Lidar. \newedit{Yeong et al. \cite{fusionsurvey} provide a survey on sensor fusion in autonomous vehicles. They categorize \msf into three primary types:} high-level fusion (HLF), mid-level fusion (MLF), and low-level fusion (LLF). \newedit{These MSFs} differ in how the data from different sensors are combined.
In HLF, each sensor independently carries out object detection or a tracking algorithm. The fusion is then conducted on the high-level object attributes of the environment (e.g., the relative positions of nearby vehicles) provided by each sensor 
and outputs aggregate object attributes to its downstream components.
LLF fuses the sensor data at the lowest level of abstraction (raw data)\cite{llf20}.
MLF is an abstraction-level between HLF and LLF. It fuses features extracted from the sensor data, such as color information from images or location features of radar and LiDAR, and then conducts recognition and classification on them\cite{mlf15}.
Among them, HLF is widely used in open-sourced commercial-grade \adas \cite{openpilot} \newedit{and}\sout{or} \ads \cite{apollo, autoware} because of its simplicity. Thus, it is the focus of the current work. 
In particular, we conduct a \carla simulator-based case study on an industry-grade \adas, \op, which uses an HLF for camera and radar.

\subsection{Fusion in \op}
\label{ssec:fusion_op}

\begin{figure}[ht]
\centering
    {\includegraphics[width=0.4\textwidth]{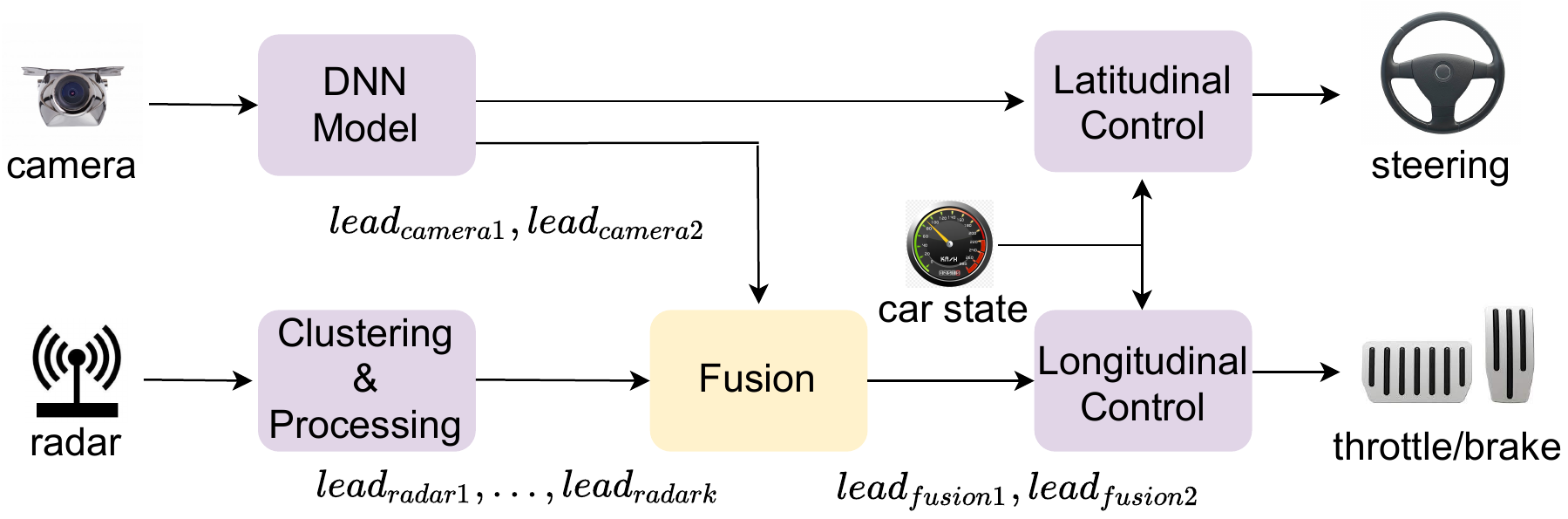}}
\caption{\small{The role of fusion in \op.}}
\label{fig:fusion_in_op}
\end{figure}

\Cref{fig:fusion_in_op} shows the the fusion component in \op. It receives data about the leading vehicles from the camera processing component and the radar processing component. Each leading vehicle data, denoted as {\it \lead}, consists of the relative speed, \sout{relative} longitudinal\newedit{,} and latitudinal distances to the leading vehicle, and the \newedit{prediction's }confidence\sout{ of this prediction} (only for camera). The fusion component aggregates all \lead information from the upstream sensor processing modules and outputs an estimation to the longitudinal control component. Finally, the longitudinal control component outputs the decisions for throttle and brake to control the vehicle. \newedit{Since }\sout{Note that }the latitudinal control component only relies on camera data\newedit{,}\sout{ so} we do not consider accidents due to the \newedit{ego car}\sout{auto-driving car} driving out of the lane. 
{Different fusion logics can be implemented. Here we studied \op default one and a popular Kalman-Filter-based fusion method \cite{fusionmathwork, attackmathworkkalmanfilter}}.

\begin{figure}[ht]
\centering
    \subfloat[\original\label{fig:fusion_logic_in_op}]
    {\includegraphics[width=0.22\textwidth]{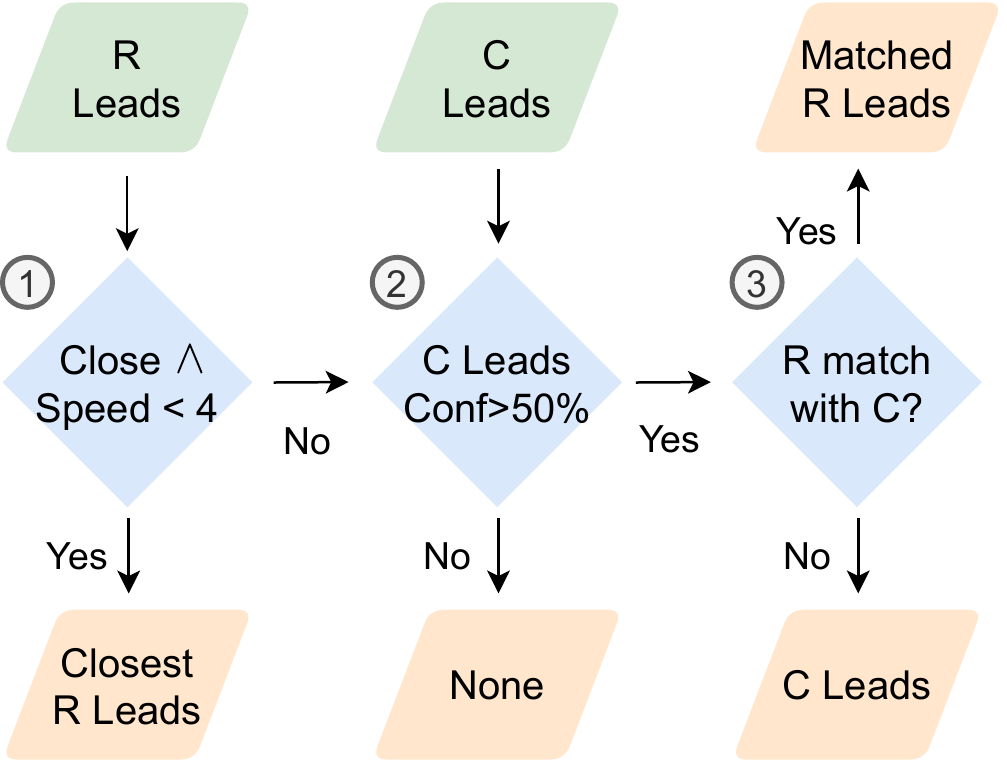}\label{fig:fusion_logic_original}}
    \hspace{5mm}
    \subfloat[\mathworkoriginal\label{fig:fusion_logic_in_mathwork}]
    {\includegraphics[width=0.22\textwidth]{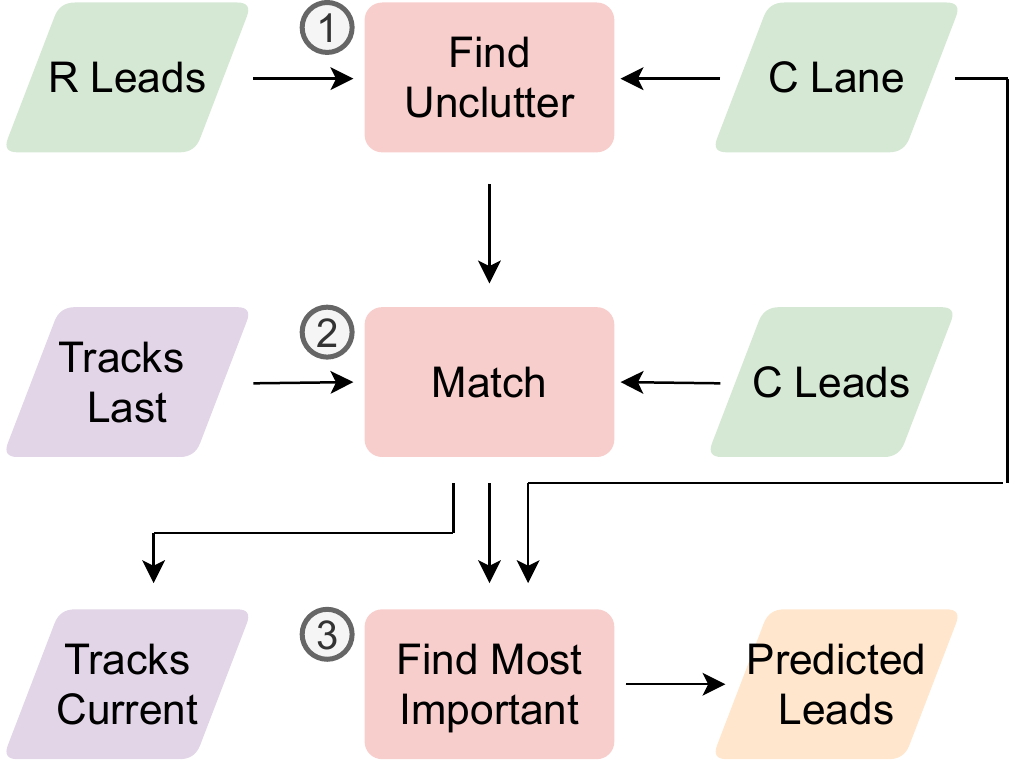}\label{fig:fusion_logic_mathworkoriginal}}
\caption{\small{Fusion logic of (a)\op \original and (b)\mathworkoriginal. C denotes camera and R denotes radar. Green, orange, blue, red, and purple denote input, output, decision, processing, and stored data over generations, respectively.}}
\label{fig:fusion_logic}
\end{figure}

\noindent\textit{\original: Heuristic Rule-based Fusion.}
\Cref{fig:fusion_logic_in_op} shows the logic flow of the \op's fusion method \original. It first checks if the ego car's speed is {\em low} (ego$_{speed}$ < 4)  and {\em close} to any leading vehicle (\textcircled{1}). 
If so, the closest radar leads are returned. 
Otherwise, it checks if the {\em confidence} of any camera leads go beyond 50\% (\textcircled{2}). If not, leading vehicles will be considered non-existent. Otherwise, it checks if any radar leads match the camera leads (\textcircled{3}). If so, the best-matching radar leads are returned. Otherwise, the camera leads are returned. 

\noindent\textit{\mathworkoriginal: Kalman-Filter Based Fusion.}
\Cref{fig:fusion_logic_in_mathwork} shows the logic of \mathworkoriginal which is a popular fusion method from Mathwork \cite{fusionmathwork}. It starts with the camera-predicted lane to filter out cluttered (i.e. stationary outside the ego car's lane) radar leads in \textcircled{1}. Then, it groups together camera leads and uncluttered radar leads, and matches them with tracked objects from last generation in \textcircled{2}. Tracked objects are then updated.
Finally, matched tracked objects within the current lane are ranked according to their relative longitudinal distances in \textcircled{3} and the data of the closest two leads are returned.

\subsection{Fusion Error \& Motivating Example}
\newedit{
A fusion error happens at the occurrence of the following: 
\begin{enumerate}[label=\roman*.,leftmargin=*]
    \item 
    Fusion logic makes some \emph{\fusionfaults} as there is disagreement between different sensor outputs, and the underlying fusion logic trusts the incorrect one even when an alternate correct input is present, and 
    \item
    A \credit{\emph{failure}}, i.e., a critical accident, takes place due to such faulty fusion method.
\end{enumerate}
}


\begin{figure}[ht]
\centering
    \subfloat[
    $time_0$ \\ 
    rel x:\\
    camera:13.7m (conf:0.1)\\
    radar:19.2m\\
    fusion:None\\
    GT:None
    \label{fig:op_bicycle_example_1}
    ]
    {\includegraphics[width=0.14\textwidth]{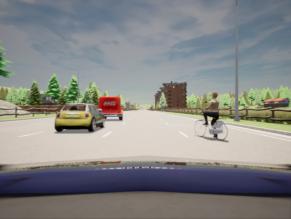}}
    \vspace{0.1mm}
    \subfloat[
    $time_1$ \\
    rel x: \\
    camera:11.8m (conf:13.5)\\
    radar: 3.9m\\
    \color{red}{fusion: None}\\
    \color{blue}{GT: 2.9m}
    \label{fig:op_bicycle_example_2}
    ]
    {\includegraphics[width=0.14\textwidth]{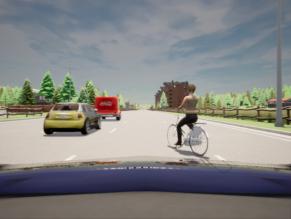}}
    \vspace{0.1mm}
    \subfloat[
    $time_2$ \\
    collision happens.
    \label{fig:op_bicycle_example_3}
    ]
    {\includegraphics[width=0.14\textwidth]{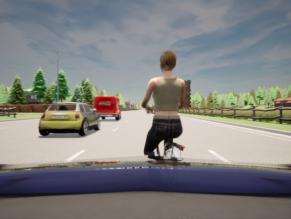}}

\caption{\small{A collision example. The ego car is the blue car \newedit{whose front view has been shown}\sout{from whose viewing angle rest of the environment can be seen}. 
{\it rel x}: relative longitudinal distance of the ego car from the {cycle}. 
{GT}: the ground-truth value.
The \newedit{\fusionfault}\sout{fusion error} and GT are highlighted in \textcolor{red}{red} and \textcolor{blue}{blue} respectively.}
}
\label{fig:motivating_example}
\end{figure} 

\noindent
\textit{Motivating Example.} \Cref{fig:motivating_example} shows an example where the ego car collides with a bicyclist cutting in. At $time_0$ (\Cref{fig:op_bicycle_example_1}), no leading vehicle exists. 
At $time_1$ (\Cref{fig:op_bicycle_example_2}), the bicyclist on the right trying to cut in. While the radar predicts that the lead is close ($3.9$ m) to the ground-truth \credit{(GT)} value (2.9m), the camera ignores the bicyclist. The fusion component trusts the camera so the ego car does not slow down, and finally a collision occurs at $time_2$(\Cref{fig:op_bicycle_example_3}). This example shows an accident caused by a wrong result from the fusion component. But how could the problem happen?

In the logic flow of the \op's  \original fusion method (see ~\Cref{fig:fusion_logic_in_op}), due to path (\textcircled{1}) and (\textcircled{2}), i.e., $\neg$(ego$_{speed}$ < 4 $\wedge$ close) and $\neg$(camera confidence > 50\%),  no leading vehicle is considered existent at $time_1$ (i.e., Fusion output=None). 
Thus, the ego car accelerates until hitting the bicyclist.

%% file: body/3_overview.tex
\section{Overview} 
\label{sec:overview}

\newedit{We focus on finding fusion errors, i.e., failures}\sout{The focus of the current paper is to find errors} caused by the faulty fusion method.
\newedit{We first define \fusionfault and \bug (see \Cref{subsec:definition}).}
Fusion methods are faulty when 
(i) two sensors' outputs differ significantly, and then (ii) the fusion logic (i.e., merging the sensor outputs) prioritizes the faulty outputs over the correct one.
To simulate \newedit{\bugs}\sout{such errors}, \tool first efficiently searches (a.k.a. fuzz) the given driving environment to find \specificss \newedit{where the fusion method tends to prioritize faulty sensor outputs}\sout{that maximize the differences between the involved sensors} and thus, lead to \newedit{failures (i.e., collisions)}\sout{accidents} (see~\Cref{subsec:fuzz}). \tool then changes the existing fusion logic with alternative ones and check whether the updated logic can avoid the  simulated crashes (see~\Cref{subsec:causality}). If a collision is avoidable with alternative fusion logic, \tool concludes that original fusion logic was erroneous.  
\new{Finally, \tool reports the unique \bugs, as described in~\Cref{sec:error_counting}.}

\sout{This section summarizes these steps.}

\subsection{Definitions}
\label{subsec:definition}

\newnewedit{We first define the \fusionfault of a fusion method $F$. 
Let, at a time step $t$, $F$ \credit{read} $m$ sensor outputs $S_{t1}$,$S_{t2}$,...,$S_{tm}$ respectively, and outputs an aggregated prediction $F_{t} := F(S_{t1},S_{t2},...,S_{tm})$. \credit{Let $GT_{t}$ denote the corresponding ground-truth value at the time step $t$.}
A correct fusion method should choose the sensor output closest to the ground truth to capture the most realistic situation.
Thus, a fusion fault occurs if there is at least one sensor input, say $S_{tj}$ at time $t$, whose distance from $GT_{t}$ is less than the distance between fusion output $F_t$ and $GT_{t}$.
To make the fault definition more tolerant to small errors, we further introduce an error tolerance threshold $th_{err}$.   
\begin{defi}
\label{fusion_fault}
$F$ makes a \textbf{\fusionfault} at a time step $t$ if
$$\min_{j\in \{1,...,m\}} \mathbf{dist}(S_{tj}, GT_{t}) + th_{err} < \mathbf{dist}(F_{t}, GT_{t}),$$ 
\end{defi}
One example of $\mathbf{dist}$ is $\mathbf{dist}(x,y)=||x-y||_1$ which is simply the l1 distance.}
\newnewedit{
Note in this work we are not interested in benign \fusionfaults that cannot lead to critical consequences. 
Besides, it is difficult to attribute a \failure to a particular \fusionfault since a \failure may appear as an effect of several \fusionfaults. Thus, we associate a \failure to the underlying fusion method.
}
\newnewedit{
\begin{defi}
A \textbf{fusion error} occurs if the system under test using the fusion method fails due to faulty fusion method.
\label{fusion_error}
\end{defi}
}

\newedit{
In this paper, we focus on the crashes of the ADAS to study fusion errors.
As per its definition, a \bug has the following two properties:
}

\begin{itemize}[leftmargin=*]
    \item \newedit{{\em Failure-inducing}: A simulation should witness a failure of the system. In our context, the system is \op and the failure is the ego car's crash. Besides, since only the longitudinal control module in \op uses fusion, we only consider ego car's collision happening within the lane it follows.}
    \item \newedit{{\em Fusion-induced}: The failure should be caused by the used fusion method. In other words, if the rest of the system and environment behave as it is, and we had a correct implementation of the fusion method, the failure would not be observed.} 
\end{itemize}

\begin{figure}[ht]
\centering
    {\includegraphics[width=0.48\textwidth]{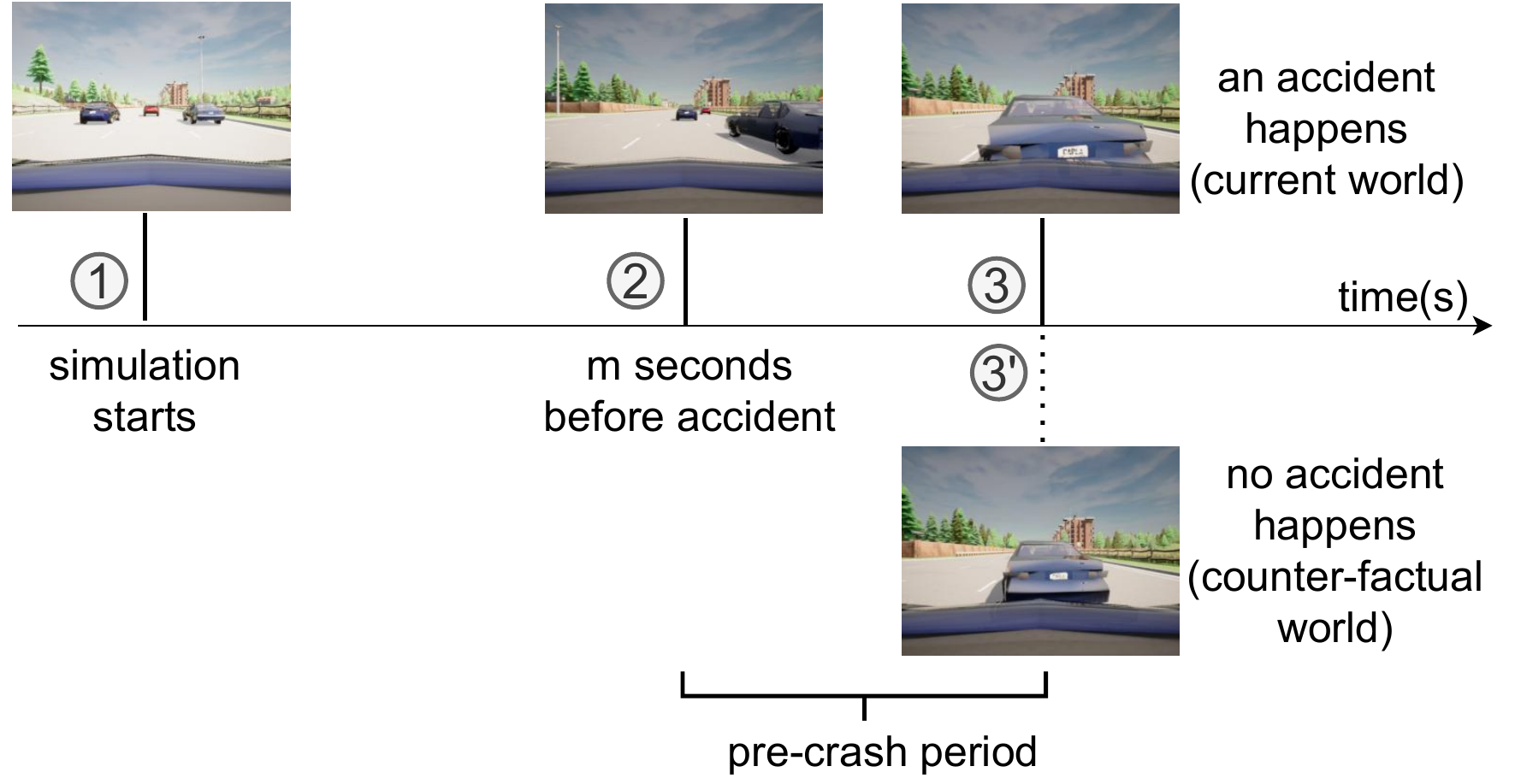}}
\caption{\small{An illustration of pre-crash period and \bug.}}
\label{fig:errordefinition_of_fusionfuzz}
\vspace{-4mm}
\end{figure}


\Cref{fig:errordefinition_of_fusionfuzz} illustrates such a \bug. At time \textcircled{1}, a simulation starts and \op is engaged. The simulation enters the pre-crash period (i.e., the $m$ seconds before an accident \new{during which it starts to misbehave}) at time \textcircled{2} and finally, a collision happens at \textcircled{3}. If \new{a better fusion method} is used from time \textcircled{2} onward, in the counter-factual world, no collision happens (at time \textcircled{~~3'~~}).

\subsection{Simulating Collisions with Fusion Fuzzing}
\label{subsec:fuzz}

To simulate \bugs efficiently, we apply an evolutionary-based fuzzing algorithm that searches for the scenarios in which \bugs are likely to happen. 
Since \bugs have two properties: \newedit{failure-inducing}\sout{critical} and fusion-induced (\Cref{subsec:definition}), we design the objective functions \newedit{to optimize }accordingly\sout{ that our fuzzing aims to optimize}. 


For capturing \newedit{{\it failure-inducing}}\sout{{\it critical} (i.e., crash-inducing)} property, we adopt the safety potential objective used in \cite{avfuzzer}. It represents the distance between the ego car and the leading vehicle (subtracted by the ego car's minimum stopping distance)---minimizing it will facilitate the collision. We denote it as $\textbf{F}_\textrm{\newedit{d}\sout{dist}}(x)$ for the scenario $x$. To further promote collision, we introduce another boolean objective function ($\textbf{F}_{\textrm{\newedit{failure}\sout{error}}}$) that is true only if a collision happens.


For the {\it fusion-induced} property, \newedit{we define an objective $\textbf{F}_{\textrm{fusion}}$ measuring \fusionfaults during an simulation, and maximize it. There can be many ways to define $\textbf{F}_{\textrm{fusion}}$. Here, we use}
\sout{we want to increase the time during which the fusion method makes an inaccurate prediction while it has received an accurate prediction from at least one sensor. To achieve this, we maximize }the number of time steps such that at each time step the fusion's output is far from the ground-truth and at least one sensor output is close to the ground-truth. Given that we use a simulated environment for testing, we can easily get the ground-truth \lead information from the simulator.
\sout{We denote this objective as $\textbf{F}_{\textrm{fusion}}$ and presents its}\newedit{We present the} details \newedit{of $\textbf{F}_{\textrm{fusion}}$} in \Cref{sec:fuzzing_algorithm}. Putting the above objectives together, we obtain the following fitness function that our evolutionary fuzzer tries to optimize (here $c_i$s are \newedit{coefficients}\sout{constants and can be set experimentally}):
\begin{equation}
    \textbf{F}(x) = c_{\textrm{\newedit{failure}\sout{error}}} \textbf{F}_\textrm{\newedit{failure}\sout{error}}(x) + c_\textrm{d} \textbf{F}_\textrm{d}(x) + c_{\textrm{fusion}} \textbf{F}_{\textrm{fusion}}(x)
\end{equation}

\subsection{Analyzing Root Causes of the Collisions} 
\label{subsec:causality}

We next analyze the simulated \newedit{\failures (i.e., collisions)}\sout{errors} reported in previous step and check 
they are indeed {\em caused} by the incorrect fusion logic. 
The most intuitive approach to check this would be to simply replace the fusion method with another fusion method and check if the collision still happens. However, this approach has two issues. 
First, compared with the initial simulation, some unobserved {influential} factors (e.g., \sout{non-determinism in sensor inputs, }the communication delay between the simulator and \op) might have {changed.}
As a result, even if a collision does not occur with an alternative fusion logic, it might be due to {the influence of} other unobserved influential factors. 
Second,  the alternative fusion logic chosen randomly may not be able to avoid the collision. Since simulation is costly, it is not possible to explore all the different logic (e.g., all the if-else branches in the fusion logic implementation~\Cref{fig:fusion_logic}-a). Thus, we must choose the alternative fusion method carefully.



To address the first issue, we resort to the theory of causal analysis. In particular, we consider the fusion method used as the interested variable and the occurrence of a collision as the interested event. We then consider all other factors that can directly or indirectly influence the collision as well as their interactions based on \newnewedit{domain knowledge, the understanding of the source code of \op and the \carla simulator, and simulation runtime behavior across multiple runs}. The goal is to control all the factors that influence the collision and are not influenced by the fusion method to stay the same across the simulations. For those influential variables that cannot be controlled directly, we apply interventions on other variables such that the uncontrollable variable's influence on the collision is eliminated. For example, to eliminate the influence of the communication latency, \newnewedit{which has been observed as the major uncontrollable influential variable, }we set the communication configurations for \op and simulator to be synchronous and deterministic. \credit{Assuming}\sout{Given the assumption} that all the influential variables are controlled, if the collision is avoided after the replacement in a counterfactual world, we can say the fusion method used is the actual cause.

To address the second issue, we define a fusion method called {\em best-sensor fusion}, which always selects the sensor prediction that is closest to the ground-truth \newedit{as per $\mathbf{dist}$ in \Cref{fusion_fault}}. This fusion method provides the best prediction among the sensors. 
Consequently, it is reasonable to assume that it should help to avoid the collision if the collision was due to the fusion method used.


\subsection{Counting Fusion Errors}
\label{sec:error_counting}
We design the principles for counting distinct \bugs in this section. Note that error counting in simulation-based testing remains an open challenge. Related works \cite{paracosm, testing_vision18} consider two errors being different if the \specificss are different. This definition tends to over-count similar errors when the search space is high-dimensional. 
Another approach manually judges errors with human efforts \cite{avfuzzer}. Such way is subjective and time-consuming when the number of errors grows up. 

Inspired by the location trajectory coverage\sout{ metric} \cite{covavfuzz}, we consider the ego car's state (i.e., location and speed) during the simulation rather than the input space variables or human judgement. 
We split the pieces of the lane that ego car drives on into $s$ intervals and the ego car's allowed speed range into $l$ intervals to get a two-dimensional coverage plane with dimensions $\mathbf{1}^{s\times l}$. During the simulation, the ego car's real-time location and speed are recorded. 
The recorded location-speed data points are then mapped to their corresponding "bins" on the coverage plane. Given all the data points mapped into the bins having the same road interval, their average speed is taken, the corresponding speed-road bin is considered "covered", and the corresponding field on the coverage plane is set $1$. Note a simulation's final trajectory representation can have at most $s$ non-zero fields. 
We denote the trajectory vector associated with the simulation run for a specification $x$ to be $\mathbf{R}(x)$ and define:
\begin{defi}
Two \bugs for the simulations runs on specifications $x_1$ and $x_2$ are considered distinct if $\vert \vert \mathbf{R}(x_1)-\mathbf{R}(x_2)\vert \vert_0 > 0$.
\label{errors_count}
\end{defi}

\begin{figure}[ht]
\centering
    \subfloat[\label{fig:traj_357}]
    {\includegraphics[width=0.034\textwidth]{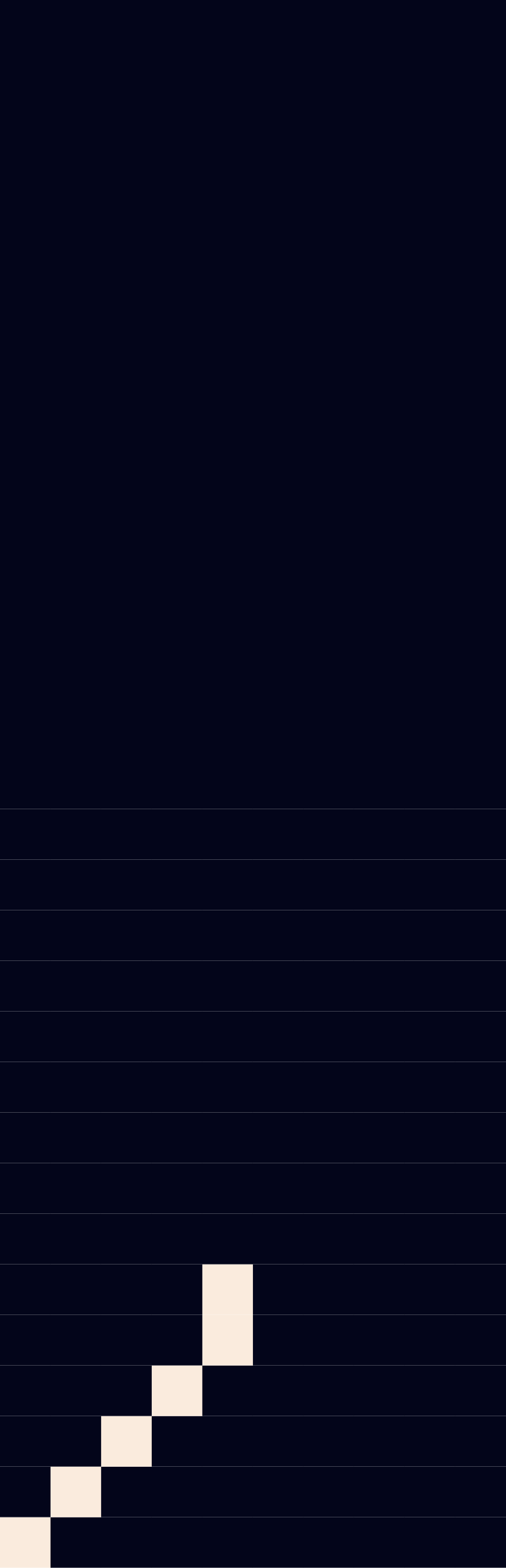}}
    \vspace{0.2mm}
    \subfloat[$time_0$]
    {\includegraphics[width=0.14\textwidth]{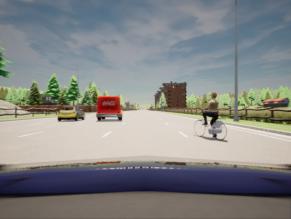}}
    \vspace{0.1mm}
    \subfloat[$time_1$]
    {\includegraphics[width=0.14\textwidth]{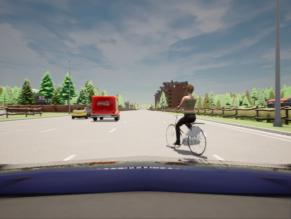}}
    \vspace{0.1mm}
    \subfloat[$time_2$]
    {\includegraphics[width=0.14\textwidth]{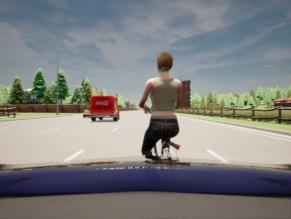}}

    \subfloat[\label{fig:traj_445}]
    {\includegraphics[width=0.034\textwidth]{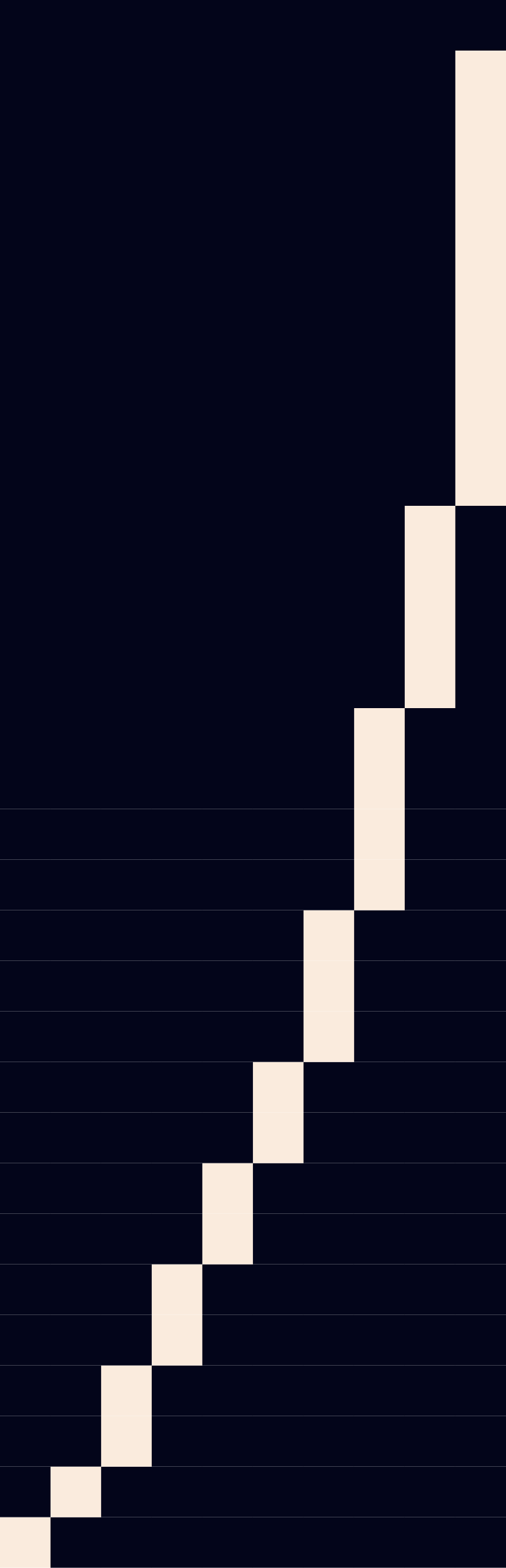}}
    \vspace{0.2mm}
    \subfloat[$time_0$]
    {\includegraphics[width=0.14\textwidth]{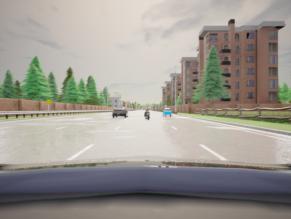}}
    \vspace{0.1mm}
    \subfloat[$time_1$]
    {\includegraphics[width=0.14\textwidth]{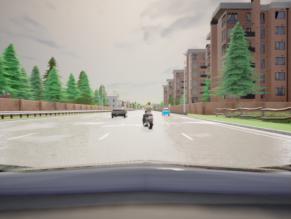}}
    \vspace{0.1mm}
    \subfloat[$time_2$]
    {\includegraphics[width=0.14\textwidth]{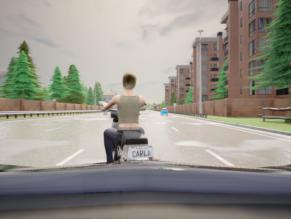}}
\caption{\small{Examples of two \bugs with different trajectories. (a) and (e) show speed-location coverage where the x-axis is speed and y axis is the road interval.}}
\label{fig:traj_demo}
\end{figure}

To demonstrate this error counting approach, we show two \bugs 
with different trajectories in \Cref{fig:traj_demo}. 
In both \Cref{fig:motivating_example} and the first row of \Cref{fig:traj_demo}, the ego car hits a bicyclist cutting in from the right lane. The difference is only that the yellow car on the left lane has different behaviors across the two runs. However, the yellow car does not influence the ego car's behavior. Hence, the two simulation runs have the same trajectory coverage (ref. \Cref{fig:traj_357}). 
By contrast, the other \bug on the second row of \Cref{fig:traj_demo} has a different trajectory (ref. \Cref{fig:traj_445}) that the ego car in high speed collides with a motorcycle at a location close to the destination. This example illustrates the necessity of counting \bugs upon \Cref{errors_count}.

%% file: body/4_fuzzing.tex
\section{\tool Methodology}
\label{sec:overview}



\begin{figure}
\centering
    {\includegraphics[width=0.48\textwidth]{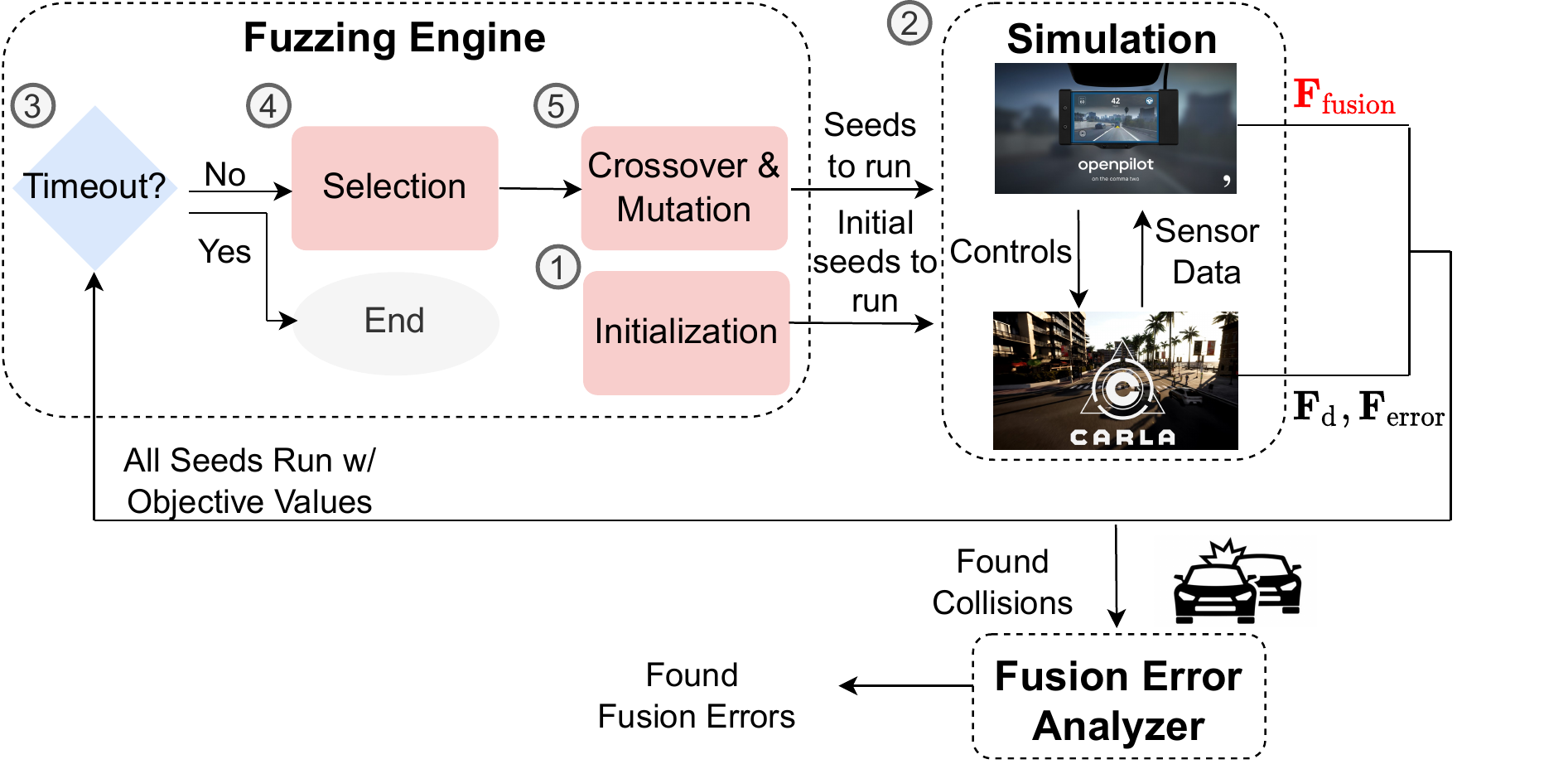}}
\caption{\small{The overall workflow of \tool.}}
\label{fig:workflow_of_fusionfuzz}
\vspace{-5mm}
\end{figure} 


In this section, we introduce \tool, our automated framework for \bugs detection. \Cref{fig:workflow_of_fusionfuzz} shows \newedit{a} high-level workflow of \tool. It consists of three major components: the fuzzing engine, the simulation, and the fusion error analyzer. The fuzzer runs for predefined rounds of generations. At each generation, it feeds generated \specificss (a.k.a. \emph{seeds}) into the simulation. In a simulation, at each time step, the \carla simulator supplies the sensor data of the current scene to \op. After \op sends back its control commands, the scene in \carla updates.
{After the simulations for all the seeds at the current generation have been run, the seeds} along with their objective values in the simulations are returned as feedback to the fuzzer. {Besides, all the collision scenarios are recorded.} 
The fuzzer then leverages the feedback to generate new seeds in the execution of the next generation. After the fuzzing process ends, all the collision scenarios are rerun with the best-sensor fusion in the counterfactual world. The scenarios that avoid the collision are reported as \bugs.

\subsection{Fuzzing Algorithm}
\label{sec:fuzzing_algorithm}
\tool aims at maximizing the number of found \bugs within a given time budget. 
\new{The search space is high-dimensional and the simulation execution is costly. Evolutionary-based search algorithms have been shown effective in such situation\cite{autofuzz21, avfuzzer}.}
We adopt an evolutionary-based search algorithm with a domain-specific fitness function (defined in \Cref{subsec:fuzz}) promoting \bugs finding. We denote our method as \alg.

The fuzzer tries to minimize a fitness function over generations. At the beginning, random seeds are sampled from the search space and fed into the simulation, as shown by \textcircled{1} in \Cref{fig:workflow_of_fusionfuzz}. In \textcircled{2}, the simulation then runs \op in \carla with the supplied scenarios. The violations found are recorded and the seeds with the objective values are returned to the fuzzer accordingly. If the whole execution runs timeout, the fuzzing procedure ends (\textcircled{3}). Otherwise, seeds are ranked based on their objective values for further {\em selection} (\textcircled{4}). The fuzzer performs {\em crossover and mutation} operations among the selected seeds to generate new seeds (\textcircled{5}) for the simulation. The steps \textcircled{2}-\textcircled{5} repeat until reaching the time threshold. \sout{The details for each step can be found in \Cref{sec:fusion_fuzz_details}.}

We next provide details for the selection step, the crossover \& mutation step, and $F_{\textrm{fusion}}$\newedit{, which is the objective promoting the occurrence of more \fusionfaults (see \Cref{subsec:fuzz})}.

\noindent\newedit{\textbf{Selection.} We use binary tournament selection, which has shown effectiveness in previous \adas testing works~\cite{testing_vision18}. For each parent candidate seed, the selection method creates two duplicates and randomly pairs up all the parent candidate seed duplicates. Each pair's winner is chosen based on their fitness function values. The winners are then randomly paired up to serve as the selected parents for the following crossover step.}

\noindent\newedit{\textbf{Crossover \& Mutation.} We adopt the simulated binary crossover \cite{SBX} following the approach in \cite{testing_vision18}. We set the distribution index $\eta=5$ and probability=$0.8$ to promote diversity of the offspring. Further, we apply polynomial mutation to each discrete and continuous variable with mutation rate set to $\frac{5}{k}$, where $k$ is the number of variables per instance, and the mutation magnitude $\eta_m=5$ to promote larger mutations.}

\noindent\newedit{\textbf{Details of $\textbf{F}_{\textrm{fusion}}$.} $\textbf{F}_{\textrm{fusion}}$ is defined as the percentage of the number of frames in which the fusion predicted \lead having a large deviation from the ground-truth \lead, while at least one predicted \lead from upstream sensor processing modules is close to the ground-truth \lead.}

\noindent\newedit{\textit{Metric Function.} In order to quantify "a large deviation" and "close to", we first define the metric function
$\mathbf{dist}(\cdot): \mathbb{R}^{|\mathbf{D}|}\times \mathbb{R}^{|\mathbf{D}|} \rightarrow \mathbb{R}$ to measure the difference between two \credit{\leads}, where $\mathbf{D}$ is a set of indices of the different dimensions of a \lead as defined in \Cref{ssec:fusion_op} (i.e., relative longitudinal distance, relative latitudinal distance, and relative speed). It takes in the two \lead and outputs their distance. The metric function can be set to any reasonable metric, e.g., l1 norm of the two \lead's difference. In the current work,
}
\newedit{
\begin{equation}
\mathbf{dist}(\hat{y}, y) := \sum_{j\in \mathbf{D}} \mathds{1}[|\hat{y}_{j}-y_{j}| > th_j],
\end{equation}
}\credit{where $\mathds{1}[\cdot]$ is a mapping of a condition's truth value to a numeric value in $\{0,1\}$.} \newedit{\credit{The function $\mathbf{dist}(\cdot,\cdot)$ thus} counts the number of dimensions of the two \credit{\leads ($\hat{y}$ and $y$) that} differ more than a corresponding error threshold $th_j$.
The threshold $th_j$ is set to $4$m, $1$m, and $2.5$m/s for relative longitudinal distance, relative latitudinal distance, and relative speed, respectively. They are chosen based on domain knowledge. In particular, the length and width of a vehicle are roughly 4m and 1.5m and the default step size of \op's cruise speed is roughly 2.5m/s. We next formally define $F_{\textrm{fusion}}$.}

\noindent\newedit{\textit{Definition of $F_{\textrm{fusion}}$.} Given a scenario $x$,}

\newedit{
\begin{equation}
\begin{split}
    \textbf{F}_{\textrm{fusion}}(x) := \frac{1}{|\mathbf{P}|}
    \sum_{t \in \mathbf{P}} \mathds{1}[&\exists~k \in \mathbf{K} \textrm{ s.t. } \mathbf{dist}(\hat{s}_{tk}, y_{t}) \leq th \\
    &\textrm{ and } \mathbf{dist}(\hat{f}_{t}, y_{t}) >th ],
\end{split}
\end{equation}}\newedit{where $\mathbf{P}$ is a set of indices of the frames during the pre-crash period, $\mathbf{K}$ is a set of indices of the sensors (camera and radar in our case), $\hat{s}_{tk}$ is a predicted lead by sensor $k$ at time frame $t$, $\hat{f}_t$ is the predicted \lead by the fusion component, $y_t$ is the ground-truth \lead, and $th$ is a distance threshold. $th$ can be set to any non-negative values. In the current work, we set $th$ to $0$. This implies that for a time frame $i$ to be counted, the fusion predicted \lead $\hat{y}_i$ must violate at least one dimension and at least one sensor's predicted \lead $\hat{s}_{ik}$ must not violate any of the three dimensions.}

\subsection{Root Cause Analysis}
\label{sec:root_cause_analysis}
This step analyzes the \bugs found by the fuzzer to confirm the incorrect fusion logic causes them. 
As described in~\Cref{subsec:causality}, we leverage causal analysis to find the root cause of the \newedit{failures}\sout{errors}, and if fusion logic is not the reason behind \newedit{a failure}\sout{an error}, we filter it out.

\begin{figure}[ht]
\centering
    {\includegraphics[width=0.4\textwidth]{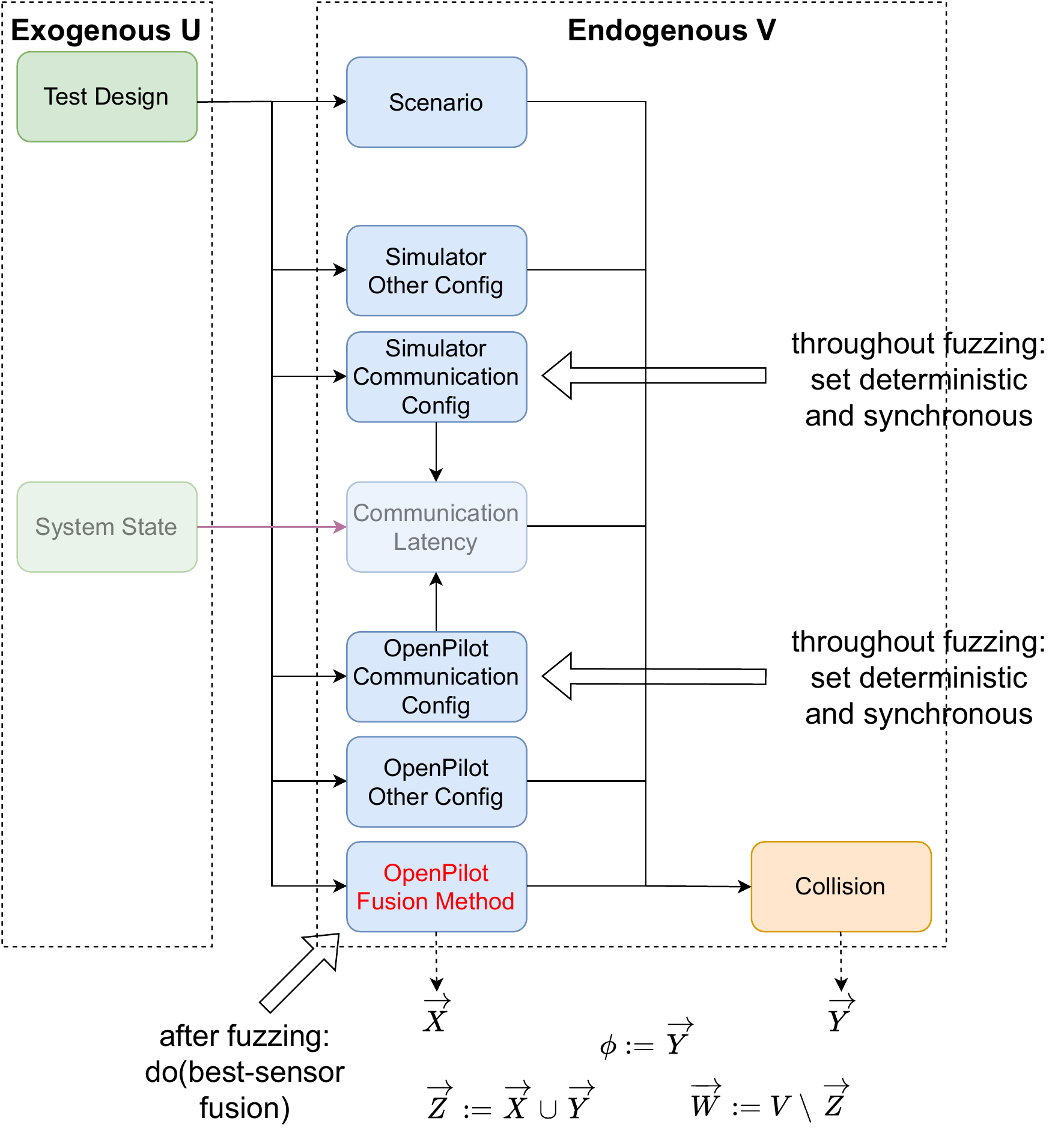}}
\caption{Illustrating the causal graph with intervention.}
\label{fig:causal_graph}
\vspace{-5mm}
\end{figure}

\paragraph{\textbf{Problem Formulation.}}
In causality analysis, the world is described by variables in the system and their causal dependencies. 
Some variables may have a causal influence on others. 
This can be represented by a Graphical Model~\cite{pearl1998graphical}, as shown in~\Cref{fig:causal_graph}, where the graph nodes represent the variables, and the edges connect the nodes that are causally linked with each other. 
For example, the test scenario should influence the occurrence of a collision. In a scenario involving many NPC vehicles, \op is more likely to \newedit{crash}\sout{misbehave}.
The variables are typically split into two sets: \emph{the exogenous variables} ($U$), whose values are determined by factors outside the model, and \emph{the endogenous variables} ($V$), whose values are ultimately determined by the exogenous variables. 

In our context, we define $\overrightarrow{X}$ to be the fusion method, $\overrightarrow{Y}$ to be a boolean variable representing the occurrence of a collision, and $\phi=\overrightarrow{Y}$. 
$\overrightarrow{Z}$ is the union of $\overrightarrow{X}$ and $\overrightarrow{Y}$. $\overrightarrow{W}$ is the complement of $\overrightarrow{Z}$ in $V$. \newedit{Following the definition of actual cause in \cite{causalitydef15},} 
\begin{defi}
\newedit{Given}\sout{Since} we know a collision (\newedit{$\phi=True$}) happens when a fusion method is used ($\overrightarrow{X}=\overrightarrow{x}$), \sout{following the actual cause of \cite{causalitydef15},}
the fusion method is an \textbf{actual cause} of \newedit{a}\sout{the} collision if: when another fusion method is used ($\overrightarrow{X}=\overrightarrow{x'}$), and all other endogenous variables (which influence the collision and are not influenced by the fusion method) are kept the same as in the original collision scenario ($\overrightarrow{W}=\overrightarrow{w}$), the collision can be avoided \newedit{($\phi=False$)}. 
\end{defi}
The details \newedit{of the justification for this definition} can be found in 
Appendix A\extend{ in the extended version \cite{fusedarxiv22}}. \newedit{We use this definition as the basis to check if a found collision is a \bug (i.e., if the used fusion method is the actual cause of the collision). In order to use it in practice, we need to (i) construct the relevant causal graph and make sure the other endogenous variables ($\overrightarrow{W}$) can be controlled, and (ii) find an alternative fusion method ($\overrightarrow{x'}$) which is likely to avoid the original collision.}

\noindent
\paragraph{\textbf{Causal Relations Analysis.}}
\sout{In order to check if a \bug satisfies this definition, we}\newedit{We} construct a causal graph (\Cref{fig:causal_graph}) specifying the relevant variables based on \newnewedit{domain knowledge, the understanding of the source code of \op and the \carla simulator, and simulation runtime behavior across multiple runs}\sout{domain knowledge}. \sout{We also make an assumption that all the variables potentially influencing the occurrence of a collision have been included in the graph.} 
The exogenous variables include test design and the state of the system running simulation (e.g., real-time CPU workload, memory usage, etc.). 
\newnewedit{Based on the understanding of \adas scenario-based testing (see \Cref{sec:related_work}), test}\sout{Test} design \newnewedit{influences the simulation result indirectly through determining}\sout{determines} scenario to test, simulator configurations, and \op configurations (including the fusion method). \newnewedit{Based on the understanding of the source code, simulator configurations can be further split into communication configurations and other configurations. Similarly, \op configurations can be split into fusion method, communication configurations, and other configurations.} \newnewedit{The other exogenous variable system state indirectly influences the collision result via an endogenous variable communication latency. This is based on our observation that, in a system with limited CPU capacity available, the latency of the sensor information passed from the simulator to \op can become very high and influences the collision result.} Communication latency collectively represents the real-time latency of the communications between the simulator and \op as well as among each of their sub-components\sout{. This variable captures the uncertainty caused by the state of the system running the simulation. It is}\newedit{, and thus captures the influence of}\sout{ influenced by} the communication configurations of simulator and \op, as well as the system state. \sout{In a system with limited CPU capacity available, the latency can become very high and results in the delay of the sensor information passed from the simulator to \op.} \sout{Consequently, collision might be more likely to happen. Finally, scenario, simulator configurations, \op configurations, and the communication latency jointly influence the occurrence of a collision in the simulation.}\newedit{We assume that all the variables directly influencing the occurrence of a collision have been included in the graph.}

\noindent\newnewedit{\textit{Intervention for Eliminating Uncontrollable Influential Variable. }}To check for causality, we need to be able to control the endogenous variables $\overrightarrow{W}$ and block any influence of the unobserved exogenous variables on the collision. \newnewedit{With the default simulator and \op communication configurations, communication latency (both between and within each of the simulator and \op) influences the collision result and prevents a deterministic simulation replay.} However, we cannot control the communication latency since one of its parents -- the system state cannot be observed and controlled. To address this issue, we set the communication configurations of the simulator and the \op to be deterministic and synchronous (see 
Appendix C \extend{in the extended version \cite{fusedarxiv22} }for details). The communication latency then becomes \newedit{zero}\sout{0} thus avoiding the potential side effects~\cite{HuGZ021}. Note such change is kept throughout the entire fuzzing process. \newnewedit{We verify that no other uncontrollable influential variables on the collision results exist after this intervention in RQ1 by checking the reproducibility of the simulation results when using the same endogenous variables.}

\noindent\newedit{\textit{Intervention for Cause Analysis.} During}\sout{Further, during} the fusion error analyzing step, we replace the initial fusion method ($\overrightarrow{x}$) with another fusion method ($\overrightarrow{x'}$) and check if a collision still happen. This step is regarded as an intervention on the fuzzing method after\sout{ the} fuzzing\sout{ process}. 

\paragraph{\textbf{Fusion Replacement Analysis.}} 
The next step is to efficiently find a fusion method $x'$ avoiding collision.
The fusion method $x'$ should possess additional properties like \newedit{having no extra knowledge}\sout{being realistic} and \newedit{being} functional. \sout{Being realistic means that it}\newedit{It} should not have extra knowledge (e.g., the ground-truth of the locations of the NPC vehicles) beyond what it receives from the upstream sensor modules. Being functional means it should be good enough to enable the ego car to finish the original task. A counter-example is if the fusion method always false positively report the presence of a stationary NPC vehicle ahead and leads the ego car to stay stationary all the time. \sout{Collision can be avoided in this case since the ego car does not move at all but the original task is not completed neither.} 

\begin{figure*}[ht]
\centering
    \subfloat[True Positive\label{fig:fusion_replacement_a}]
    {\includegraphics[width=0.32\textwidth]{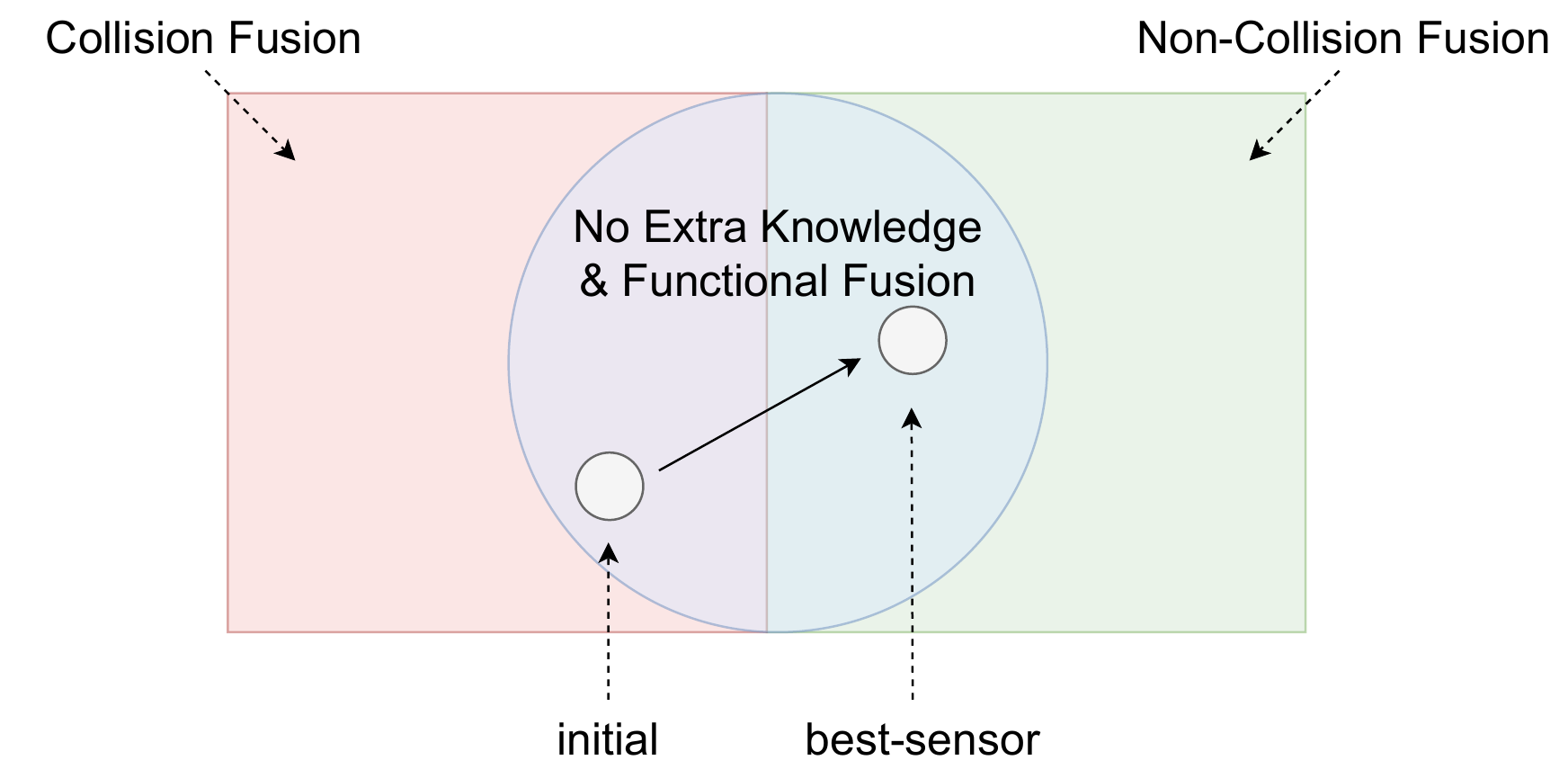}}
   ~
    \subfloat[True Negative\label{fig:fusion_replacement_b}]
    {\includegraphics[width=0.32\textwidth]{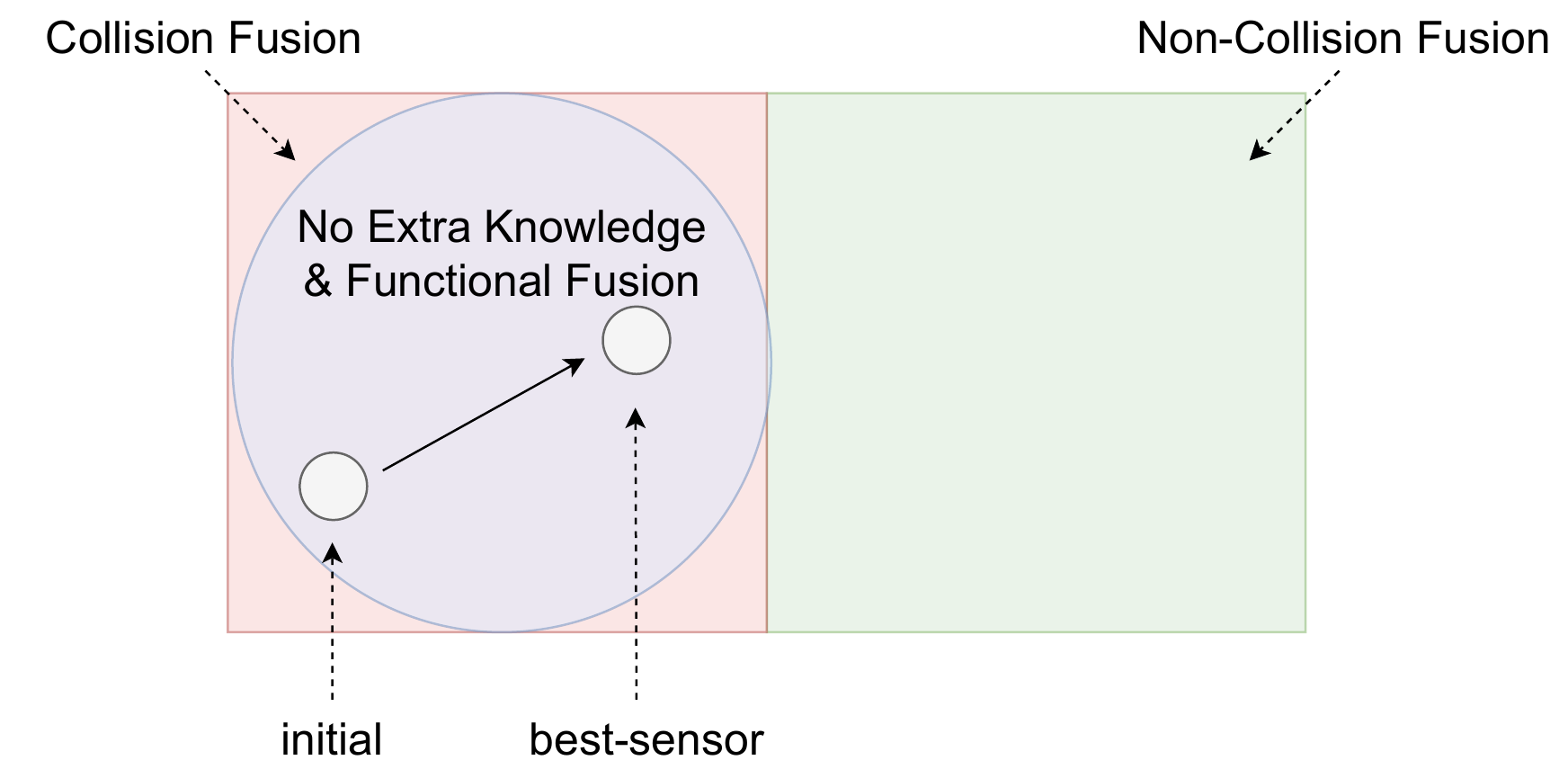}}
    ~
    \subfloat[False Negative\label{fig:fusion_replacement_c}]
    {\includegraphics[width=0.32\textwidth]{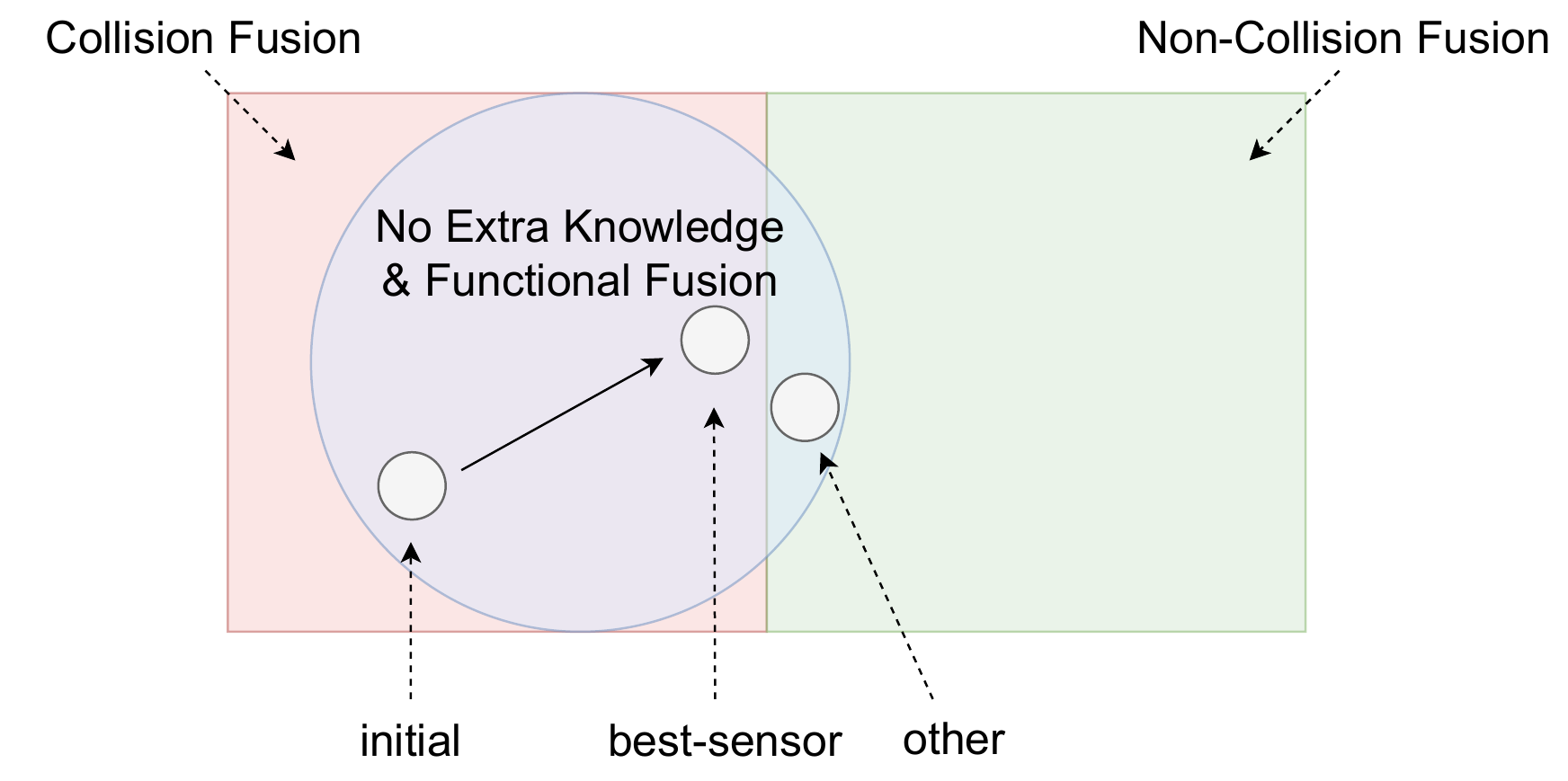}}
    
\caption{\small{An illustration of three situations of replacing the fusion method.}}
\label{fig:fusion_replacement}
\end{figure*}

To illustrate this, we define three different classes of fusion methods. Given everything else is kept the same, \emph{collision fusion class} and \emph{non-collision fusion class} consist of the fusion methods that lead to and avoid the collision, respectively. \emph{\newedit{No extra knowledge}\sout{Realistic} \& functional class} consists of fusion methods which \newedit{have no extra knowledge}\sout{are both realistic} and \newedit{are} functional. If an \newedit{failure}\sout{error} is caused by the fusion method and can be fixed by changing it to a \newedit{no extra knowledge}\sout{realistic} and functional fusion method, there should be an intersection between non-collision fusion and \newedit{no extra knowledge}\sout{realistic} \& functional fusion as shown in \Cref{fig:fusion_replacement}(a) and \Cref{fig:fusion_replacement}(c). 
Otherwise, there should be no intersection \sout{between the two and the realistic \& functional fusion class should be a subset of the collision fusion class }as shown \Cref{fig:fusion_replacement}(b). 
The initial fusion method should fall into the intersection of the collision fusion class and the \newedit{no extra knowledge}\sout{realistic} \& functional fusion class since a collision happens and it is reasonable to assume it (\original or \mathworkoriginal) \newedit{has no extra knowledge}\sout{is realistic} and \newedit{is} functional. 


In the current work, for each found collision, we only run one extra simulation to check if the fusion method is the cause. In particular, we set $\overrightarrow{x'}$ to \emph{best-sensor fusion}. Note that this fusion method is an oracle fusion method since in reality we won't be able to know the ground-truth.
However, it serves a good proxy.
\newedit{First, it uses no additional knowledge except for using the \sout{. Although for our purpose, it uses} ground-truth to select the best sensor output. In reality, an ideal fusion method might potentially select the most reliable upstream sensor's prediction even without additional knowledge.}
\sout{First, it is realistic. Although it uses ground-truth, in reality, a fusion method might potentially achieve the best-sensor fusion method's output which is the most reliable upstream sensor's prediction.} 
Second, it is functional since it provides more accurate prediction than methods like \original and thus should be able to finish the original cruising task. 
Third, if \emph{best-sensor fusion} cannot help to avoid a collision after the replacement, there is a high possibility that the collision is not due to the fusion method. The reason is that it already picks the best sensor prediction \newedit{and thus does not make \fusionfault, }and it is reasonable to assume that the downstream modules\sout{ are more likely to} perform better given its output compared with those less accurate outputs.

Thus, best-sensor fusion serves as a proxy to check if there is an intersection between \newedit{no extra knowledge}\sout{realistic} \& functional fusion and non-collision fusion. If best-sensor fusion can help avoid the collision, the \newedit{\failure}\sout{error} will be considered a \bug. Otherwise, it will be discarded.
There are three situations: (a) the \newedit{\failure}\sout{error} is caused by the fusion method and the best-sensor fusion falls into non-collision fusion class(\Cref{fig:fusion_replacement_a}). (b) the \newedit{\failure}\sout{error} is not caused by the fusion method and the best-sensor fusion does not fall into non-collision fusion class(\Cref{fig:fusion_replacement_b}). (c) the \newedit{\failure}\sout{error} is caused by the fusion method and the best-sensor fusion does not fall into non-collision fusion class(\Cref{fig:fusion_replacement_c}). (a) and (b) are the true positive and true negative cases since the causation of the fusion method is consistent with the collision results of the best-sensor fusion method, while (c) is the false negative case. It also should be noted that there is no false positive case since if best-sensor fusion helps avoiding the collision, according to our reasoning earlier, the causation must hold. The implication is that \sout{a}\newedit{a predicted }\bug \newedit{is a failure}\sout{ is an error} caused by the fusion method but the reverse does not always hold.

%% file: evaluation/rq1.tex
\section{Results}

To evaluate \tool, we explore the following research questions:

\textbf{RQ1: Evaluating Performance.} {How effectively can \tool find \bugs in comparison to baselines?}

\textbf{RQ2: Case Study of Fusion Errors.} {What are the representative causes of the \bugs found?}

\textbf{RQ3: Evaluating Repair Impact.} {How to improve \newedit{MSF}\sout{Multi-Sensor Fusion} in \op based on our observations on found \bugs?}

\input{evaluation/setup}

\subsection{RQ1: Evaluating Performance}

\begin{figure}[ht]
\centering
    \includegraphics[width=0.23\textwidth]{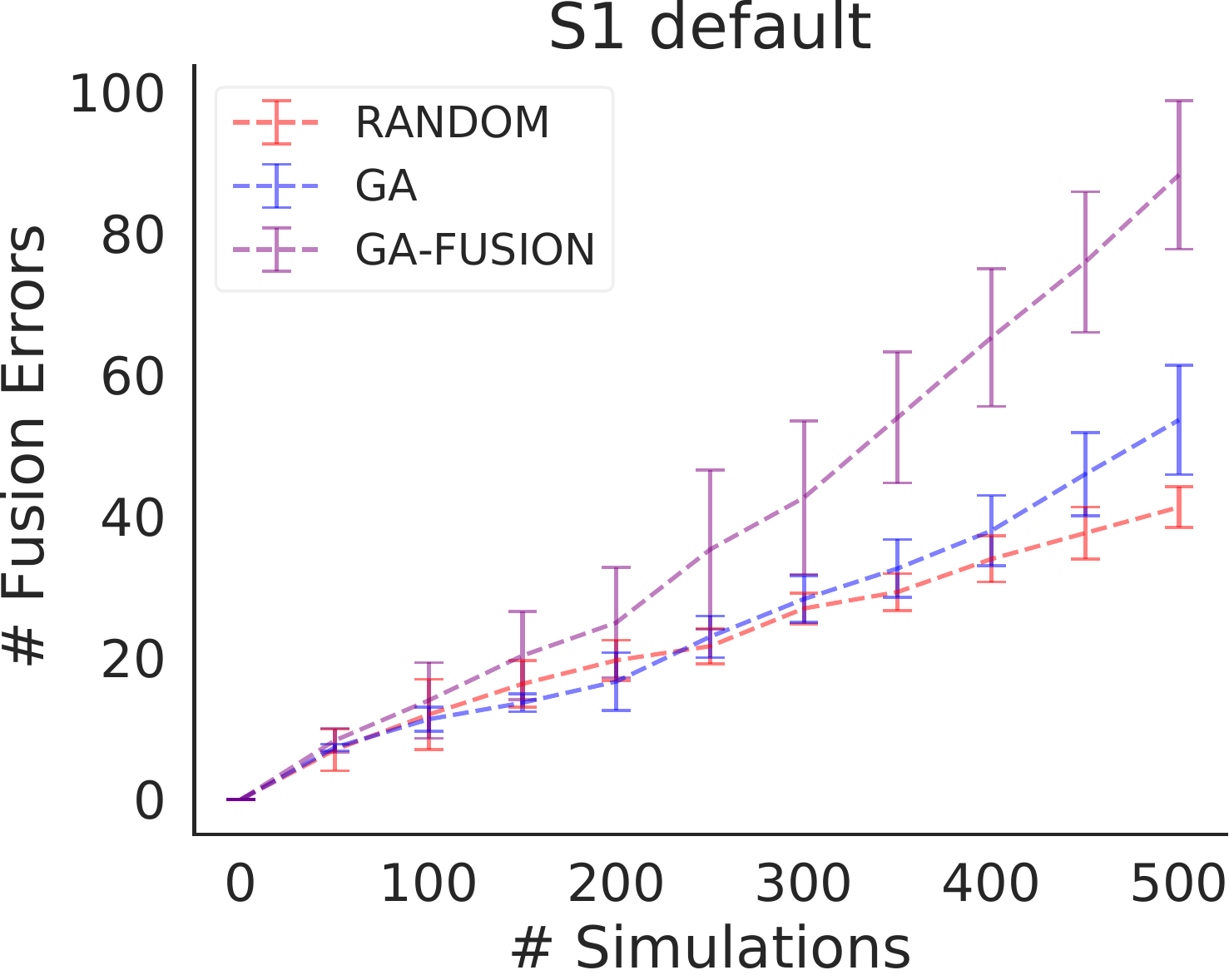}
    \includegraphics[width=0.23\textwidth]{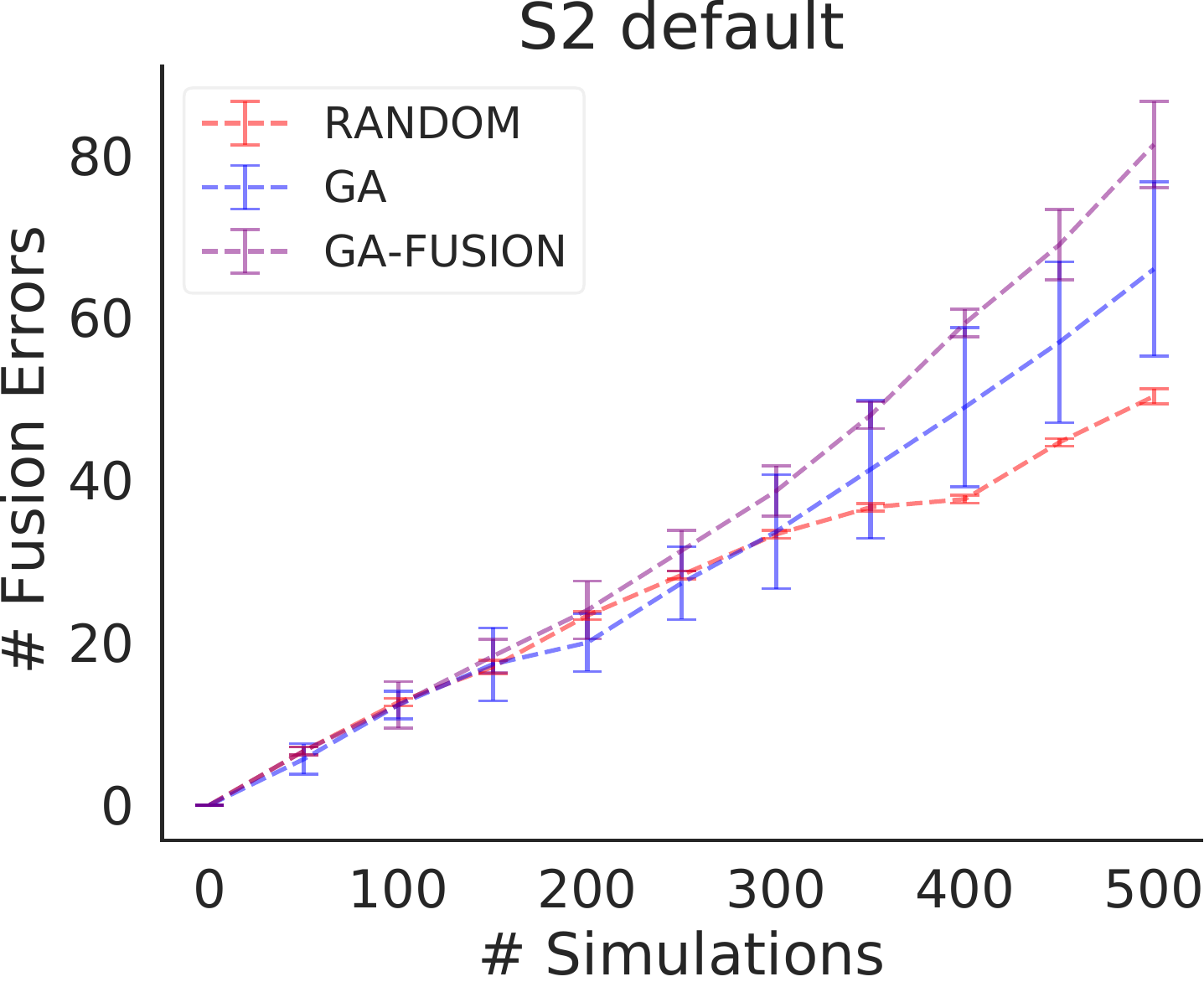}
    \includegraphics[width=0.23\textwidth]{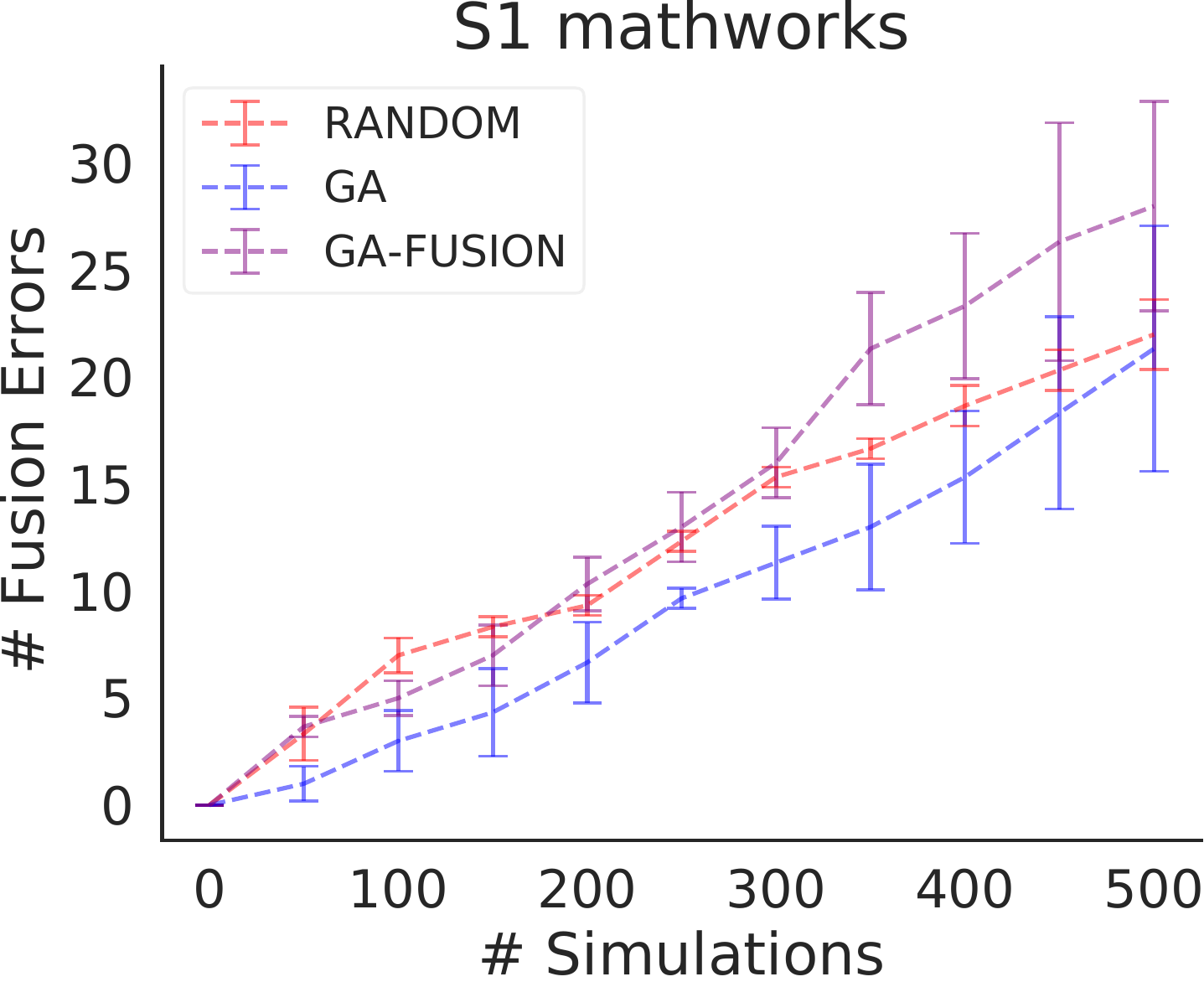}
    \includegraphics[width=0.23\textwidth]{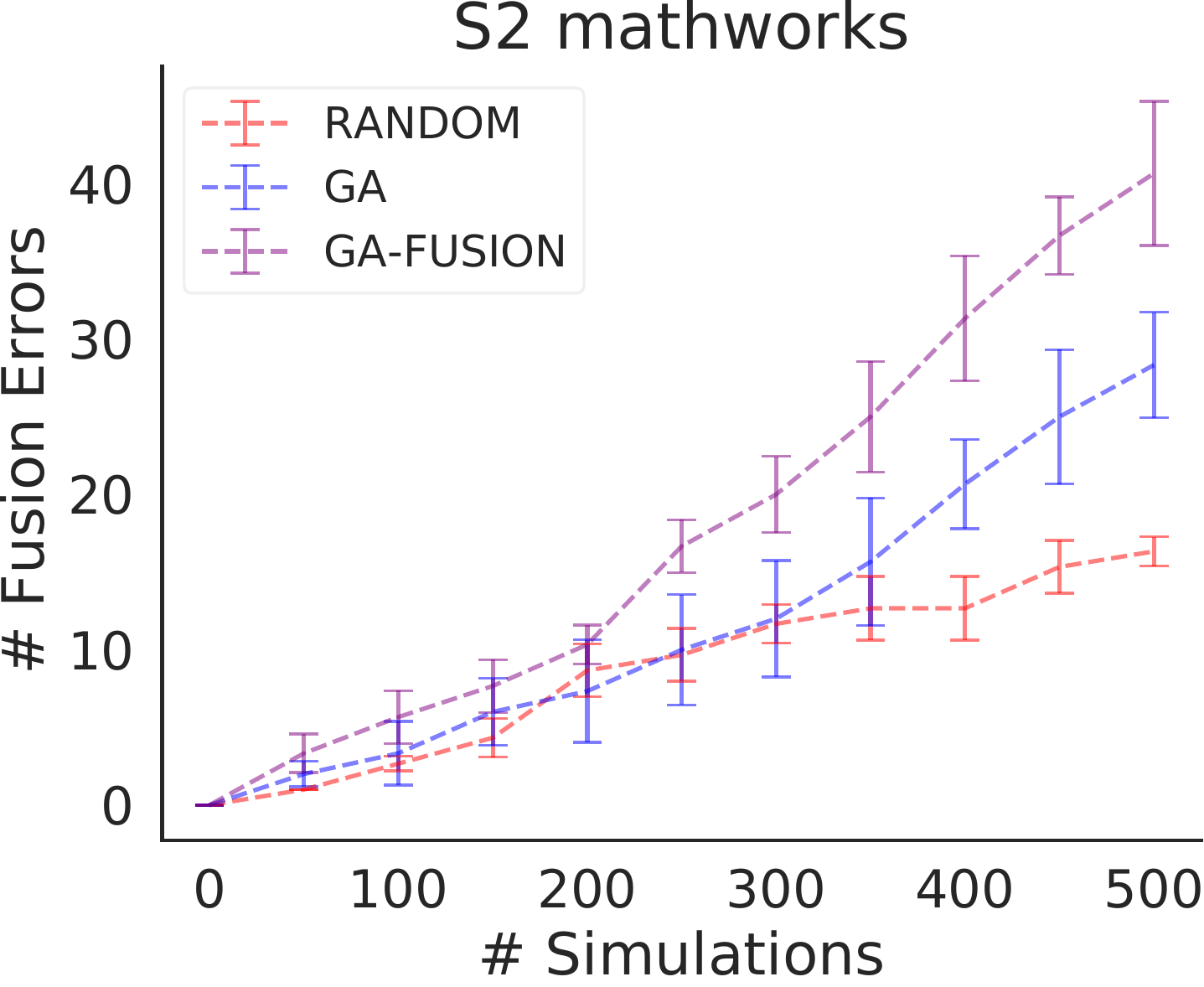}
    
\caption{\small{\# \bugs found over \# simulations.}}
\label{fig:bug_num_results_all}
\end{figure}

\begin{figure}[ht]
\centering
    \includegraphics[width=0.23\textwidth]{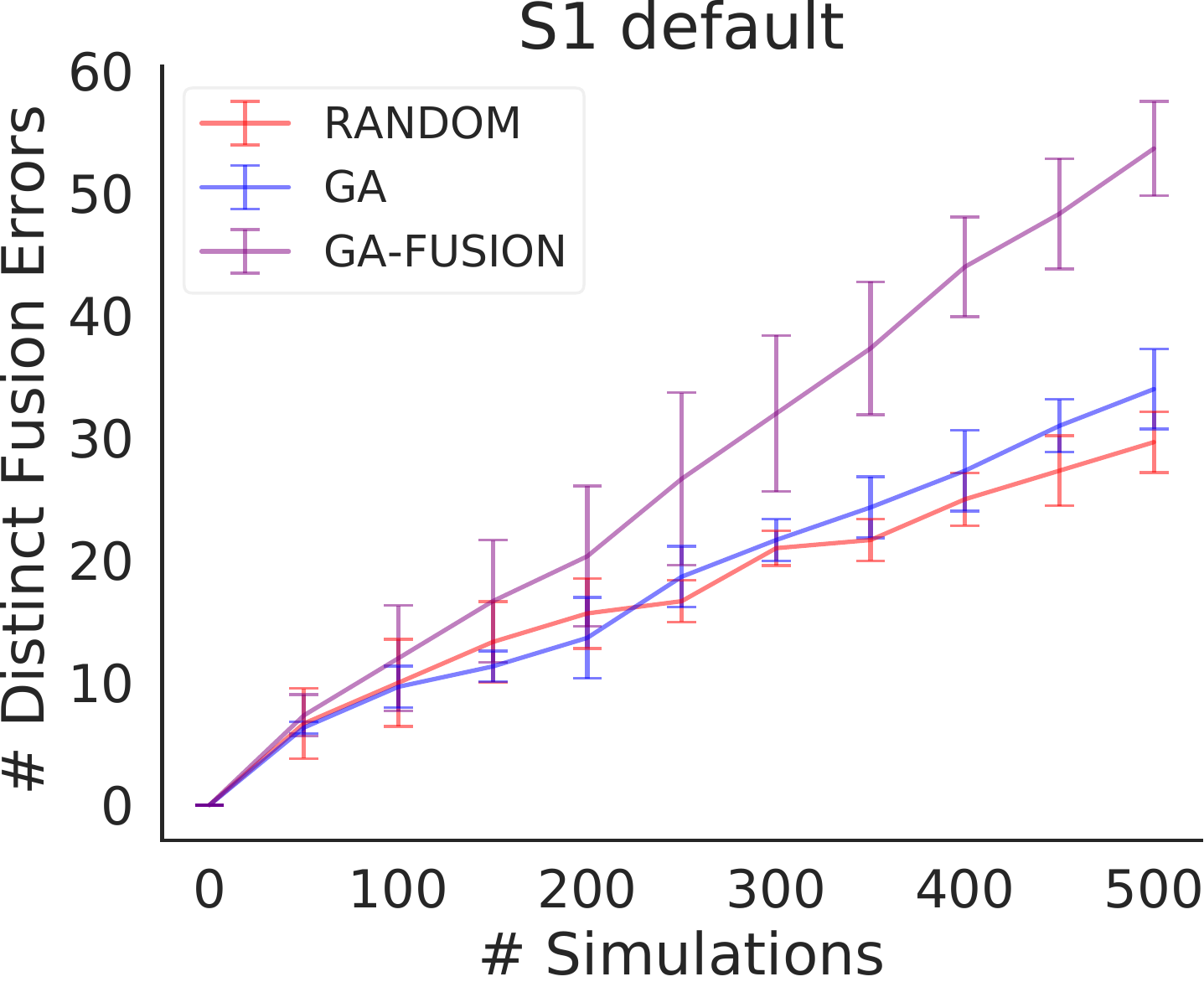}
    \includegraphics[width=0.23\textwidth]{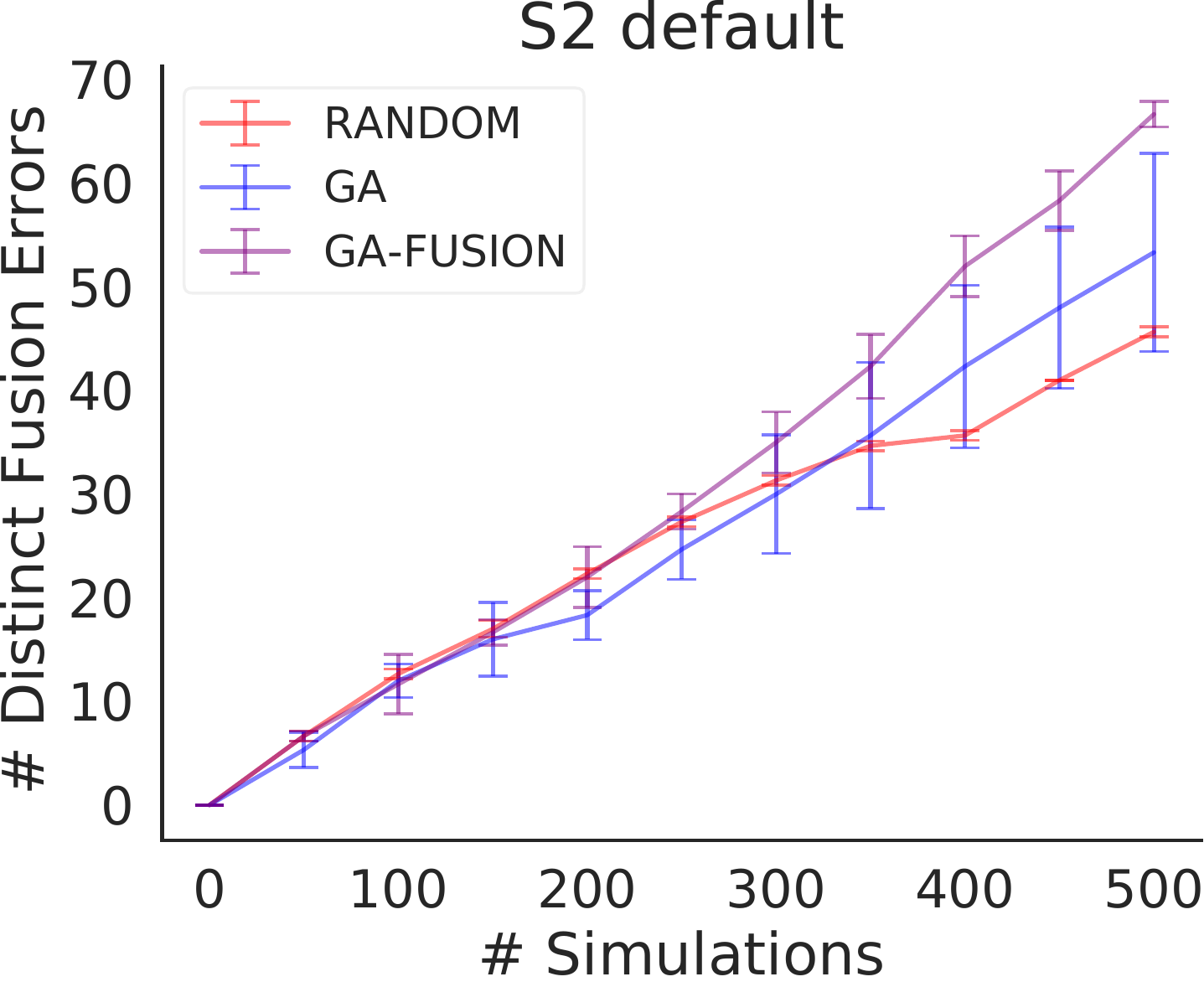}
    \includegraphics[width=0.23\textwidth]{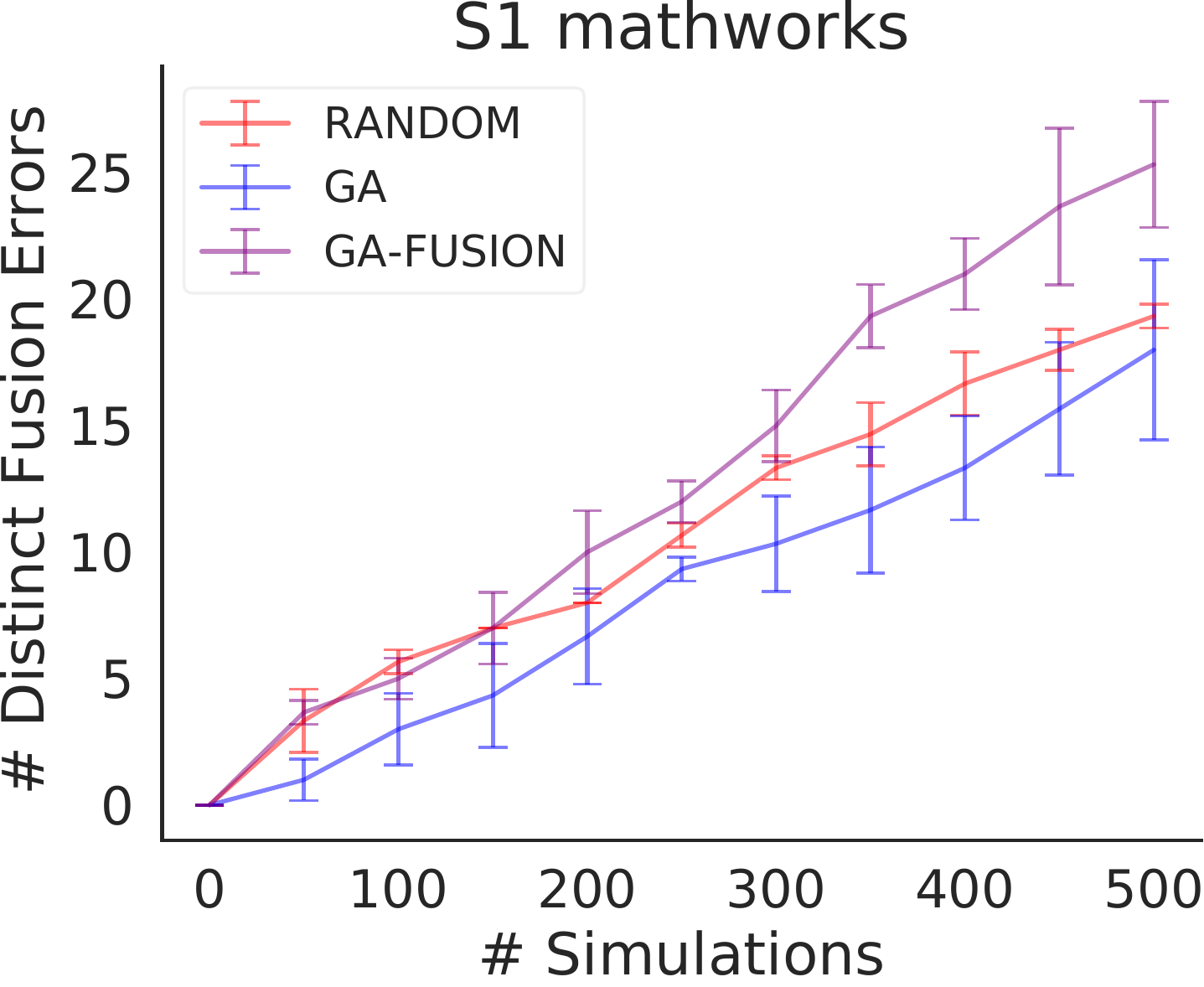}
    \includegraphics[width=0.23\textwidth]{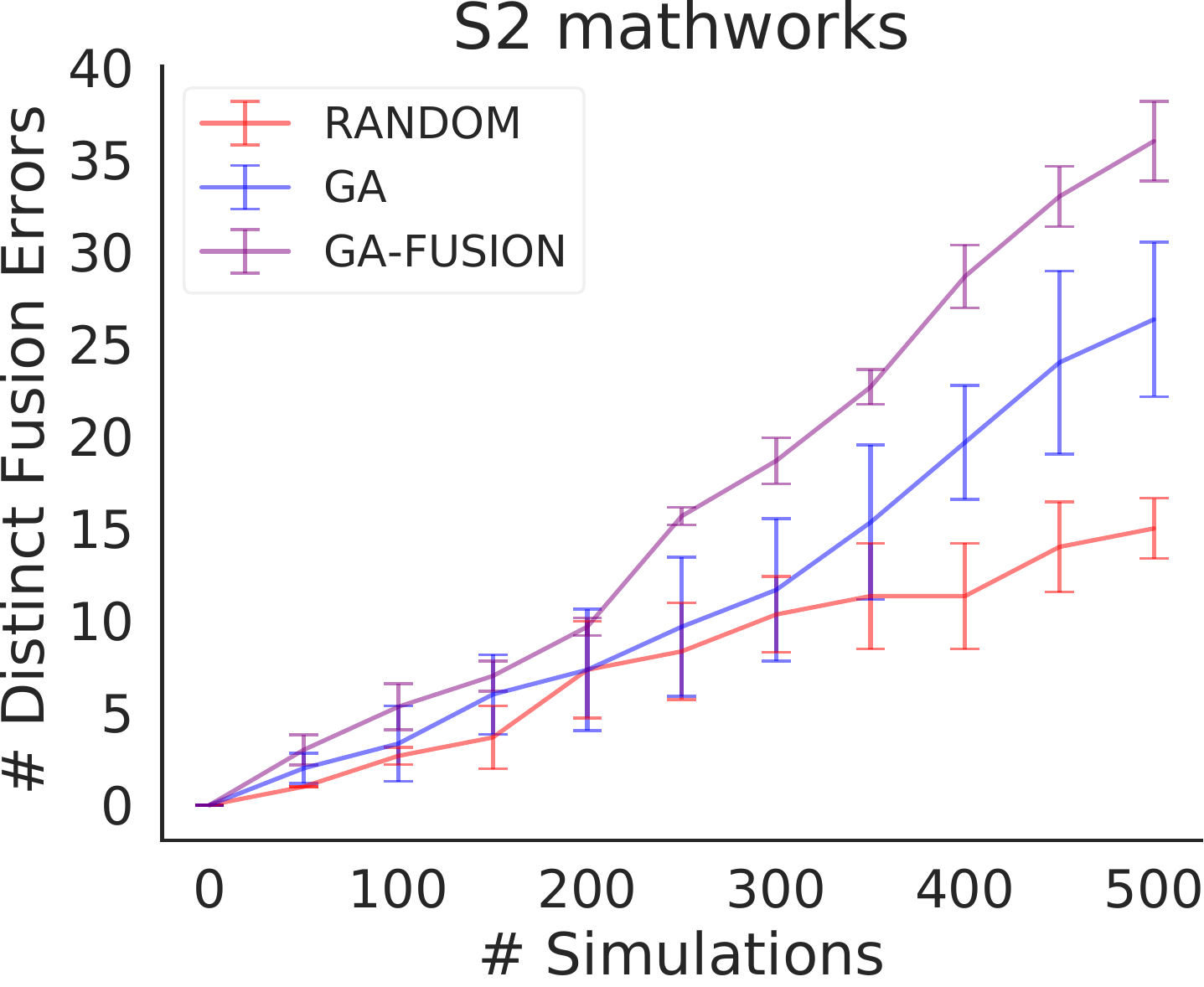}
    
\caption{\small{\# distinct \bugs found over \# simulations.}}
\label{fig:bug_num_results}
\end{figure}

We compare \alg with the two baselines. \Cref{fig:bug_num_results_all} shows the average number of \bugs found by the three methods over three runs for each setting. On average\newedit{,} \alg has found 65\%, 27\%, 23\%, \newedit{and }44\% more \bugs than the best baseline method under each setting, respectively. 
\Cref{fig:bug_num_results} shows the average number of distinct \bugs (based on \Cref{errors_count}) found by the three methods over three runs for each setting. \alg has also found 58\%, 31\%, 25\%, and 37\% more distinct \bugs 
than the best baseline method, respectively.
To test the significance of the results, we further conduct Wilcoxon rank-sum test \cite{Wilcoxon} and Vargha-Delaney effect size test \cite{VDtest, guidetostatstest} between the number of distinct \bugs found by \alg and the best baseline under each setting. For each of the four settings, we have the p-value 0.05 and VD effect size interval (0.68, 1.32) at the 90\% confidence interval, suggesting the difference is significant and the difference has medium effect size.
These results show the effectiveness of \tool and superiority of \alg. \newedit{The result also holds under different pre-crash period $m$ (see 
Appendix G\extend{ in the extended version \cite{fusedarxiv22}}.}

\newnewedit{The proposed \alg can efficiently find more \bugs because: (1) $\textbf{F}_{\textrm{d}}$ and $\textbf{F}_{\textrm{failure}}$ can differentiate collisions (including both \bugs and \nfbugs) from no-collision, and (2) $\textbf{F}_{\textrm{fusion}}$ can differentiate \bugs from \nfbugs. The first point is straightforward since when a collision happens, $\textbf{F}_{\textrm{d}}$ is usually smaller and $\textbf{F}_{\textrm{failure}}$ is $1$. To show the second point, we plot the empirical cumulative density functions (ECDFs) of $\textbf{F}_{\textrm{fusion}}$ for no-collision, \nfbugs, and \bugs, respectively, when running \alg under each setting.
}

\begin{figure}[!hpt]
\centering
    \subfloat[\original on S1\label{fig:op_fusion_perc_error_s1}]
    {\includegraphics[width=0.23\textwidth]{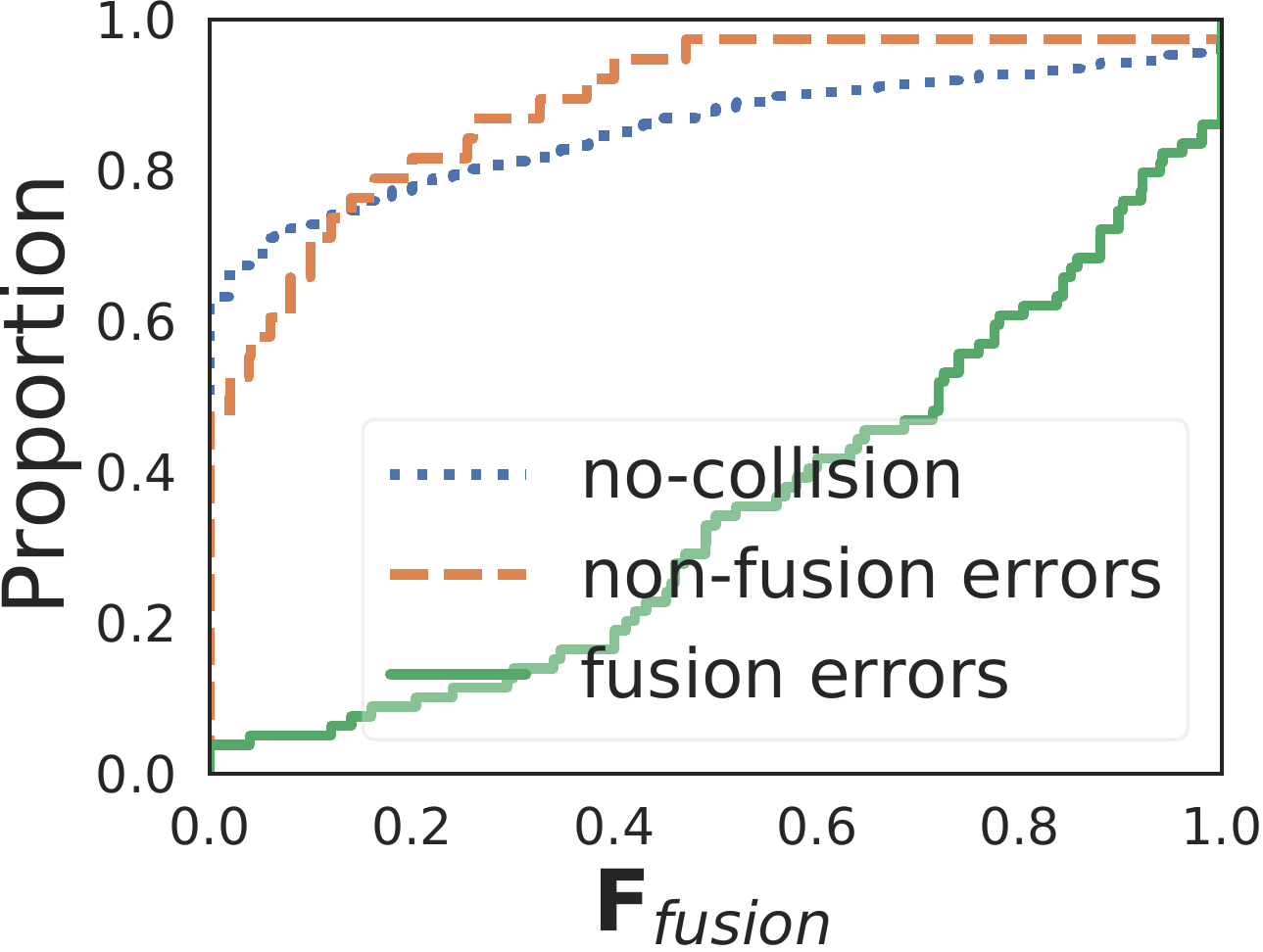}}
    \vspace{0.1mm}
    \subfloat[\mathworkoriginal on S1\label{fig:mathwork_fusion_perc_error_s1}]
    {\includegraphics[width=0.23\textwidth]{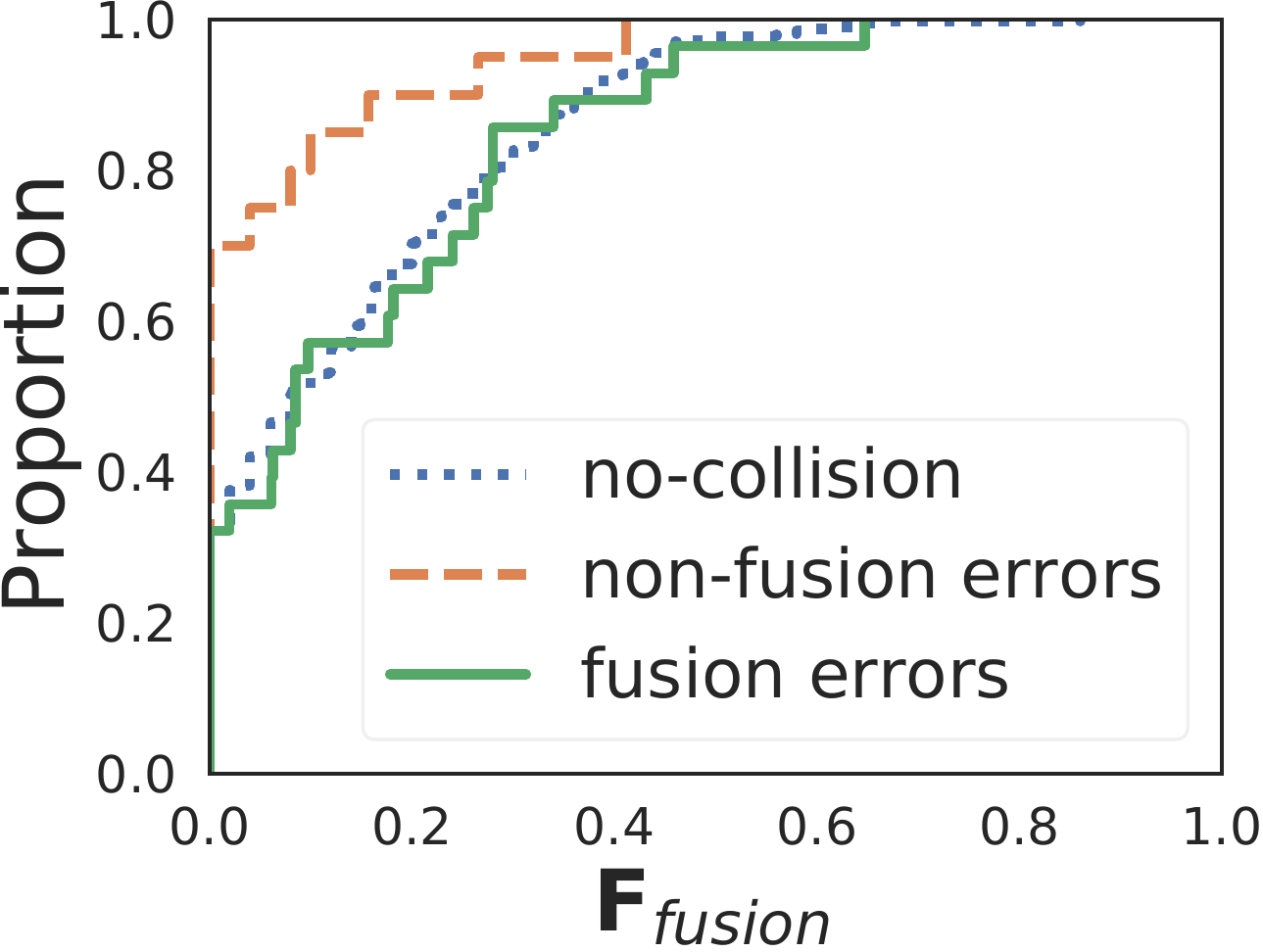}}
    \vspace{0.1mm}
    \subfloat[\original on S2\label{fig:op_fusion_perc_error_s2}]
    {\includegraphics[width=0.23\textwidth]{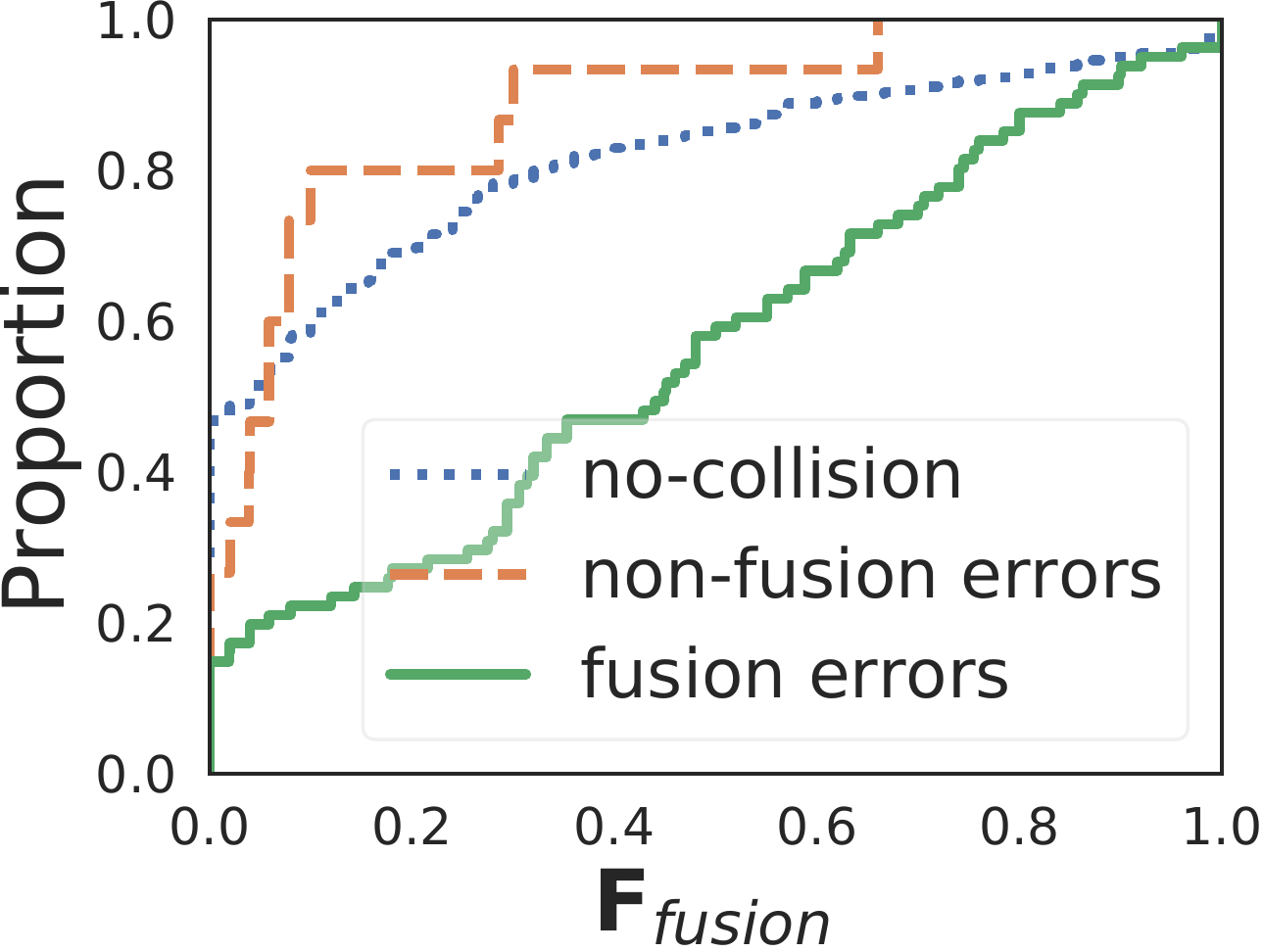}}
    \vspace{0.1mm}
    \subfloat[\mathworkoriginal on S2\label{fig:mathwork_fusion_perc_error_s2}]
    {\includegraphics[width=0.23\textwidth]{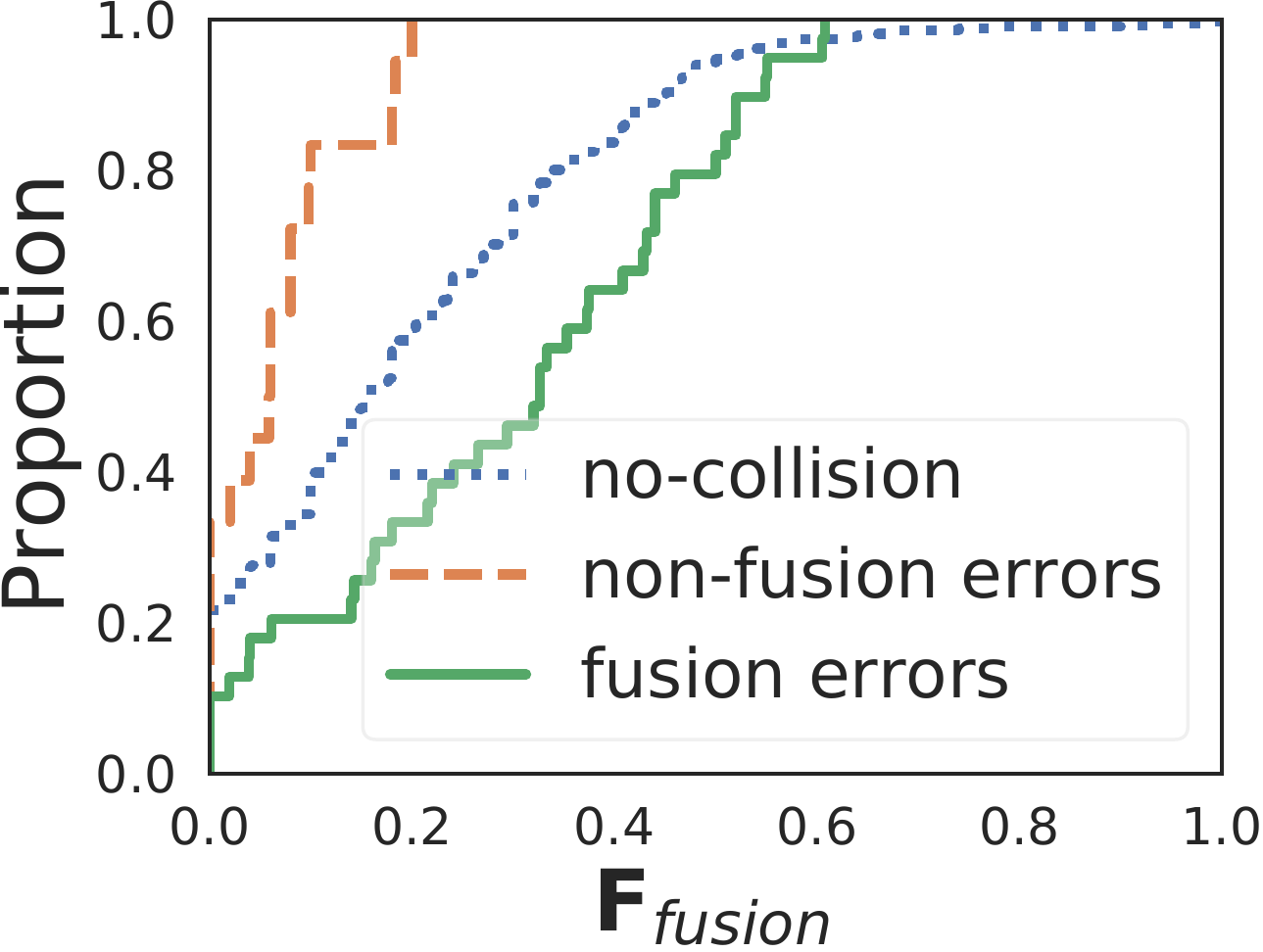}}
    \vspace{0.1mm}
\caption{\small{\newnewedit{Empirical Cumulative Density Functions (ECDFs) of $\mathbf{F}_{\textrm{fusion}}$ for the three groups under the four settings. For each group, at a given x-axis value, the y-axis value is the proportion of scenarios in the group that have $\mathbf{F}_{\textrm{fusion}}$ less than or equal to the given x-axis value. The plots show that a larger portion of \bugs have larger $\mathbf{F}_{\textrm{fusion}}$ than \nfbugs so $\mathbf{F}_{\textrm{fusion}}$ can help to differentiate the two.}}}
\label{fig:fusion_perc_correlation}
\vspace{-5mm}
\end{figure}

\newnewedit{As shown in \Cref{fig:fusion_perc_correlation}, on average, \bugs tend to have larger $\mathbf{F}_{\textrm{fusion}}$ than \nfbugs. We also apply Two-sample Kolmogorov–Smirnov test \cite{smirnov1939estimation} on the ECDFs of $\mathbf{F}_{\textrm{fusion}}$ for \bugs and \nfbugs. The test statistic versus the corresponding 0.05 significance threshold for each setting are 0.76>0.27, 0.42>0.39, 0.58>0.38, and 0.67>0.39, respectively \cite{massey1952distribution, TAOCP}, showing the \bugs and \nfbugs differ at the 0.05 significance level under each setting.
}

\newnewedit{Note that in \Cref{fig:fusion_perc_correlation}(b), \bugs have similar $\mathbf{F}_{\textrm{fusion}}$ as no-collision. 
This can happen when fusion method is faulty but that may not result in collision. Typical examples are scenarios in which a leading vehicle's relative longitudinal distance is wrongly predicted but no collision happens since it is very far away from \op.
However, these \bugs are still more likely to be selected at the selection stage since \bugs involve the occurrence of collisions which result in smaller $\textbf{F}_{\textrm{d}}$ and larger $\textbf{F}_{\textrm{failure}}$ than no-collision.
}

\noindent\newnewedit{\textbf{Sanity Check of the Causal Graph.} In order to make sure the causal graph (\Cref{fig:causal_graph}) includes all the influential variables on the collision result, from the scenarios we have run during the fuzzing process, we randomly selected 100 collision scenarios and 100 no-collision scenarios, and run them again with every controllable endogenous variable kept the same. All the repeated runs reproduce the collision/no-collision results. This implies no influential variables are likely to be omitted, since if such variables exist, repeated runs with the same endogenous variables should lead to different simulation results.}


\RS{1}{\newedit{Under each of the four settings, at the 0.05 significance level, \tool finds more distinct \bugs (as well as \bugs) than the best baseline method. The difference has a medium effect size at 90\% confidence interval.}}

%% file: evaluation/setup.tex
\subsection{Experimental Design}


\begin{figure}[ht]
\centering
    {\includegraphics[width=0.45\textwidth]{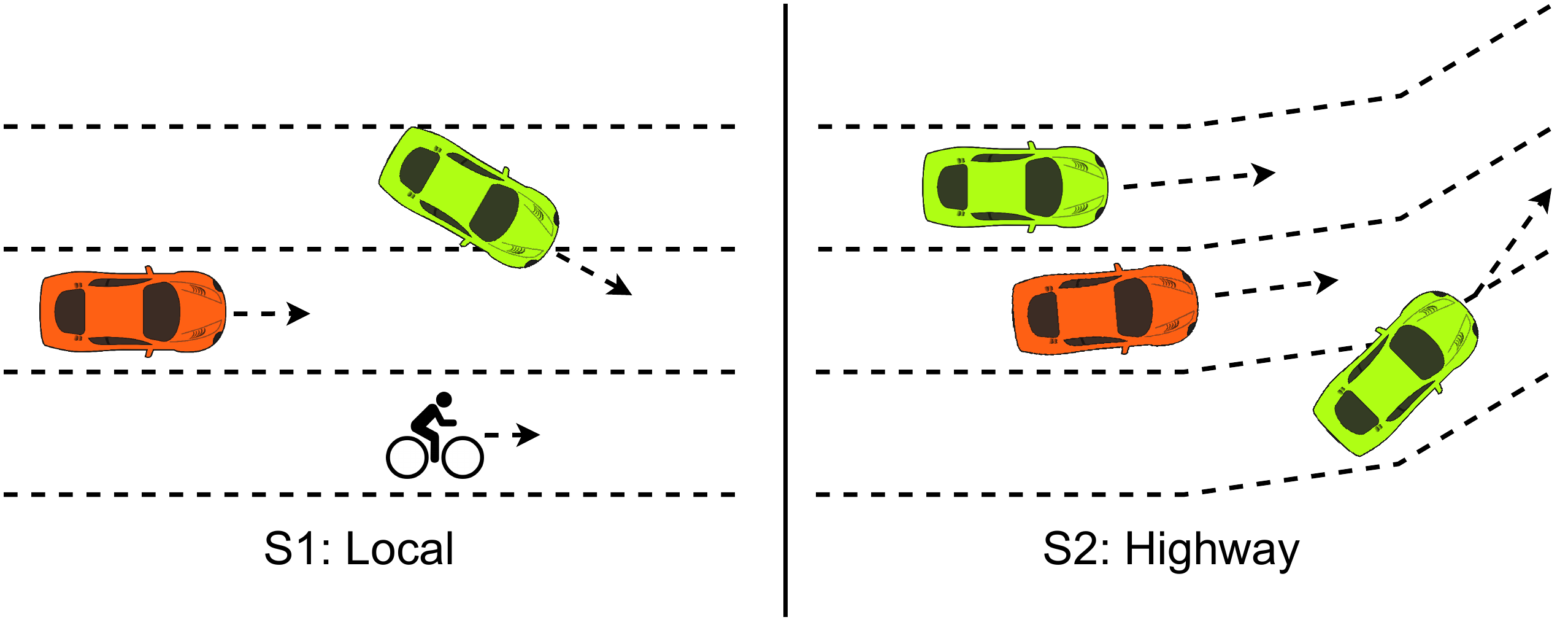}}
\caption{\small{An illustration of the two driving environments where the orange car represents the ego car.}}
\label{fig:scenarios_illustration}
\vspace{-5mm}
\end{figure}

\noindent\textbf{Environment.} We use \CARLA 0.9.11 \cite{carla} as the simulator and \op 0.8.5 as the \adas ~\cite{openpilot}. The experiments run on a Ubuntu20.04 desktop with Intel i9-7940x, Nvidia 2080Ti, and 32GB memory.

\noindent\textbf{Studied Fusion Methods.}
We apply \tool on \original and \mathworkoriginal introduced in \Cref{ssec:fusion_op}.

\noindent\textbf{Driving Environments.} 
We utilize two \lss named \Sa and \Sb\sout{ in our study}. \Sa is a straight local road and \Sb is a left curved highway road.
Both \Sa and \Sb have 6 NPC vehicles. An illustration is shown in \Cref{fig:scenarios_illustration} (not all NPC vehicles are shown).
The maximum allowed speed of the auto-driving car is set to $45$ miles/hr \newedit{($\approx 20.1$m/s)} on the highway road (\Sb) and $35$ miles/hr \newedit{($\approx 15.6$m/s)} on the local road (\Sa). For vehicle types, \Sb only considers cars and trucks while \Sa additionally includes motorcycles and bicyclists. The search space for each vehicle consists of its type, its speed\sout{ (from 0 to 150\% of the maximum allowed speed of the current road)} and lane change decision (turn left/right, or stay in lane) at each time interval. Weather and lighting conditions are also \newedit{considered}\sout{searchable}. 
\sout{In total, The search space consists of $76$ dimensions (see \Cref{sec:search_space_details} for details).}
See 
Appendix E and Appendix F\extend{ in the extended version \cite{fusedarxiv22}} for the details of the driving environments and how they comply to the capability of \op, respectively.


\noindent\textbf{Baselines and Metrics.}
We use random search (\RA) and genetic algorithm without $\textbf{F}_{\textrm{fusion}}$ in the fitness function (\GA) as two baselines. We set the number of \specificss causing \bugs and distinct \bugs (\Cref{errors_count}) as two evaluation metrics.

\noindent\textbf{Hyper-parameters.} We set the default values for $c_{\textrm{\newedit{failure}\sout{error}}}, c_{d}, c_{\textrm{fusion}}$ in the fitness function to\sout{ be} $-1,1,-2$. Since the crash-inducing property has two terms {($c_{\textrm{\newedit{failure}\sout{error}}}$ and $c_\textrm{d}$)} while the fusion aspect\sout{ only} has one ({$c_{\textrm{fusion}}$}), the\sout{ choice of these} default values balances the \newedit{two's} contribution\sout{ of the two}. The sign for $c_{\textrm{d}}$ is positive since we want to minimize $F_{\textrm{d}}$ and the signs for the other two are negative since we want to maximize $F_{\textrm{\newedit{failure}\sout{error}}}$ and $F_{\textrm{fusion}}$.
We set the pre-crash period's $m$ to 2.5 seconds because several states in US use 2.5s as the standard driver reaction time and studies have found the 95 percentile of perception-reaction time for human drivers is 2.5s \cite{reactiontime12}. Besides, in our context, when a fusion-induced collision happens, it is often caused by the fusion component’s failure for about 2.5s before the collision as in \Cref{fig:motivating_example}. We set $s$ and $l$ (\new{defined in \Cref{sec:error_counting}}) to 30 and 10 such that each road interval is about $5$m and each speed interval is about $4$m/s. By default, we fuzz for 10 generations with 50 simulations per generation; each simulation runs at most 20 simulation seconds.

%% file: evaluation/rq2.tex
\subsection{RQ2: Case Study of Fusion Errors.}
\label{sec:showcases}
In this subsection, we show three representative \bugs found by \tool and analyze their root causes.

\begin{figure}[ht]
\centering
    \subfloat[
    $time_0$, rel x:\\
    camera:12.6m\\ 
    (conf:95.4)\\
    radar:11.9m\\
    fusion:11.9m\\
    GT:10.5m 
    \label{fig:op_vehicle_collision_1}]
    {\includegraphics[width=0.11\textwidth]{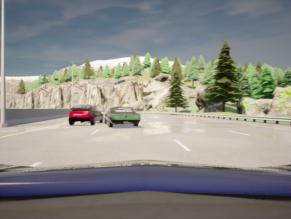}}
    \vspace{0.1mm}
    \subfloat[
    $time_1$, rel x:\\
    camera:8.9m\\
    (conf:49.9)\\
    radar:6.9m\\
    \color{red}{fusion:none}\\
    \color{blue}{GT:5.4m} 
    \label{fig:op_vehicle_collision_2}]
    {\includegraphics[width=0.11\textwidth]{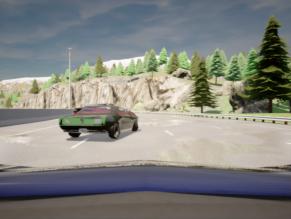}}
    \vspace{0.1mm}
    \subfloat[$time_2$,\\
    a collision \\
    happens. 
    \label{fig:op_vehicle_collision_3}]
    {\includegraphics[width=0.11\textwidth]{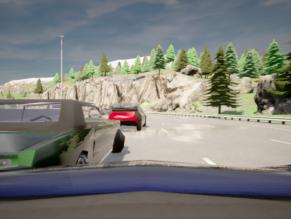}}
    \vspace{0.1mm}
    \subfloat[
    $time_2'$,\\
    no collision.
    \label{fig:op_vehicle_collision_3_best}]
    {\includegraphics[width=0.11\textwidth]{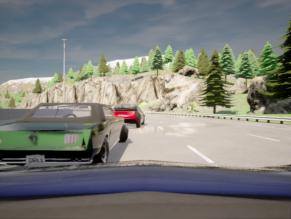}}
    
    \subfloat[
    $time_0$, rel x:\\
    camera:23.2m\\
    (conf:47.1)\\
    radar:15.5m\\
    \color{red}{fusion:none}\\
    \color{blue}{GT:14.4m}
    \label{fig:op_vehicle_collision_2_1}]
    {\includegraphics[width=0.11\textwidth]{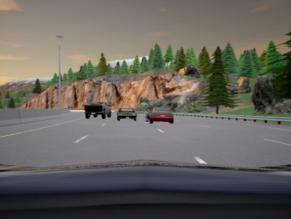}}
    \vspace{0.1mm}
    \subfloat[$time_1$, rel v: \\
    camera:-6.8m/s
    (conf:62.3)\\
    radar:-14.1m/s\\
    \color{red}{fusion:-6.8m/s}\\
    \color{blue}{GT:-14.0m/s}
    \label{fig:op_vehicle_collision_2_2}]
    {\includegraphics[width=0.11\textwidth]{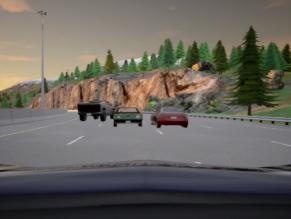}}
    \vspace{0.1mm}
    \subfloat[
    $time_2$,\\
    a collision \\
    happens.
    \label{fig:op_vehicle_collision_2_3}]
    {\includegraphics[width=0.11\textwidth]{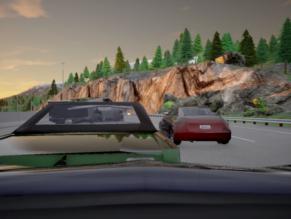}}
    \vspace{0.1mm}
    \subfloat[
    $time_2'$,\\
    no collision.
    \label{fig:op_vehicle_collision_2_3_best}]
    {\includegraphics[width=0.11\textwidth]{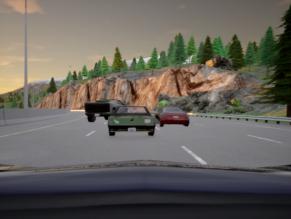}}
    
\caption{\small{Two found \bugs for \original. {\it rel v} represents relative speed to the leading NPC vehicle.}}
\label{fig:failure_demos}
\vspace{-5mm}
\end{figure}

\noindent\textbf{Case1: Incorrect camera lead dominates accurate radar lead.}
The first row of \Cref{fig:failure_demos} shows a failure due to \sout{the misbehavior of} \textcircled{2} in \Cref{fig:fusion_logic_original}. 
In \Cref{fig:op_vehicle_collision_1}, both camera and radar give accurate prediction of the leading green car at $time_0$. At $time_1$ in \Cref{fig:op_vehicle_collision_2}, the green car tries to change lane, collides with a red car, and blocks the road. The camera model predicts that the green car with a low confidence (49.9\%) and the fusion component thus misses all leading vehicles due to \textcircled{2} in \Cref{fig:fusion_logic_original}. The ego car keeps driving until hitting the green car at $time_2$ in \Cref{fig:op_vehicle_collision_3}. If the radar data is used instead from $time_0$, however, the \newedit{collision}\sout{accident} can be avoided, as shown in \Cref{fig:op_vehicle_collision_3_best}. Going back to \Cref{fig:fusion_logic_original}, the root cause is that \op prioritizes the camera prediction and it ignores any leading vehicles if the camera prediction confidence is below \new{50\%}. As a result, despite the accurate information predicted by the radar, \op still causes the collision.

\noindent\textbf{Case2: Inaccurate radar lead selected due to mismatch between radar and camera.}
The second row of \Cref{fig:failure_demos} shows another failure caused by both \textcircled{2} and \textcircled{3} in \Cref{fig:fusion_logic_original}. 
At $time_0$ of \Cref{fig:op_vehicle_collision_2_1}, camera overestimates the longitudinal distance to the leading green car. At $time_1$  of \Cref{fig:op_vehicle_collision_2_2}, though one radar data (not shown) is close to the correct information of the green car, the radar data of the Cybertruck on the left lane matches the camera's prediction. Thus, the Cybertruck lead data is selected regarding \textcircled{3} in \Cref{fig:fusion_logic_original}. Consequently, although the ego car slows down, the process takes longer time than if it selects the green car radar data. This finally results in the collision at $time_2$ in \Cref{fig:op_vehicle_collision_2_3}. If the green car radar lead is used from $time_0$, the ego car would slow down quickly and thus not hit the green car at $time_2'$ in \Cref{fig:op_vehicle_collision_2_3_best}. This failure also correlates to camera dominance but it additionally involves mismatching in \textcircled{3}
of \Cref{fig:fusion_logic_original}.

\begin{figure}[ht]
\centering
\subfloat[
    $time_0$, rel x: \\
    camera:10.3m
    (conf:86.3)\\
    radar:7.4m\\
    \color{red}{fusion:10.5m}\\
    \color{blue}{GT:7.2m}
    \label{fig:op_vehicle_collision_3_1}]
    {\includegraphics[width=0.11\textwidth]{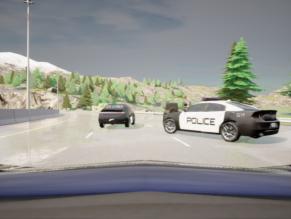}}
    \vspace{0.1mm}
    \subfloat[
    $time_1$, rel x: \\
    camera:6.8m\\
    (conf:97.8)\\
    radar:3.0m\\
    fusion:3.0m\\
    GT:2.9m
    \label{fig:op_vehicle_collision_3_2}]
    {\includegraphics[width=0.11\textwidth]{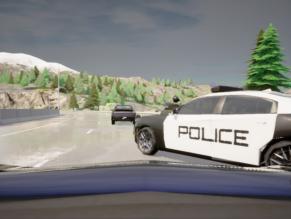}}
    \vspace{0.1mm}
    \subfloat[
    $time_2$,\\
    a collision \\
    happens.
    \label{fig:op_vehicle_collision_3_3}]
    {\includegraphics[width=0.11\textwidth]{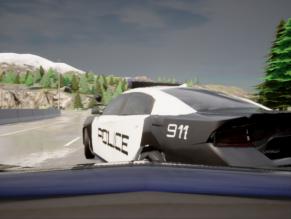}}
    \vspace{0.1mm}
    \subfloat[
    $time_2'$,\\
    no collision.
    \label{fig:op_vehicle_collision_3_3_best}]
    {\includegraphics[width=0.11\textwidth]{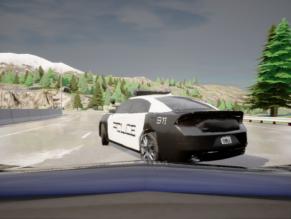}}
\caption{\small{A found \bug for \mathworkoriginal.}}
\label{fig:failure_demos_mathworkoriginal}
\vspace{-3mm}
\end{figure}

\noindent\textbf{Case3: Discarding correct lead due to a faulty selection method.}
\Cref{fig:failure_demos_mathworkoriginal} shows an example when \mathworkoriginal fails due to \textcircled{3} in \Cref{fig:fusion_logic_mathworkoriginal}. At $time_0$ in \Cref{fig:op_vehicle_collision_3_1}, a police car on the right lane is cutting in. While radar gives a very accurate prediction of the red car, the radar prediction is not used since \textcircled{3} in \Cref{fig:fusion_logic_mathworkoriginal} only selects among the predicted leads within the current lane. Consequently, a camera predicted lead is used, which overestimates the relative longitudinal distance. At $time_1$ in \Cref{fig:op_vehicle_collision_3_2}, the correct radar prediction is used but it is too late for the ego car to slow down, causing the collision at $time_2$ in \Cref{fig:op_vehicle_collision_3_3}. If the best predicted lead (i.e. the one from the radar data) is used starting at $time_0$, the collision would disappear at the time $time_2'$ of \Cref{fig:op_vehicle_collision_3_3_best}.

\RS{2}{The representative \bugs found by \tool for the two fusion methods are due to the dominance of camera over radar, their mismatch, or the faulty \newedit{prediction} selection method\sout{ of the prediction}.}

%% file: evaluation/rq3.tex
\subsection{RQ3: Evaluating Repair Impact.}

\new{To avoid the fusion errors, one obvious alternative to the studied fusion methods seems to simply let the radar predictions dominate the camera predictions as the examples shown in \Cref{sec:showcases} are mainly caused by the dominance of unreliable camera prediction. Such design, however, suffers from how to choose the fusion leads from all radar predicted leads. If the fusion method simply chooses the closest radar lead and that lead corresponds to an NPC vehicle on a neighboring lane, the ego car may never pass the NPC vehicle longitudinally even when the NPC vehicle drives at a low speed.}

\begin{table}[h]
\small
    \vspace{-2mm}
    \centering
    \caption{\small{\# avoided ~/ \# distinct \bugs. \label{tab:bug_num_results}}}
    \begin{tabular}{l|l|l|l}
    \toprule
        S1 & S1 & S2 & S2 \\
        \original & \mathworkoriginal & \original & \mathworkoriginal \\
    \toprule
       43/52 & 14/26 & 67/78 & 13/26\\
    \bottomrule
    \end{tabular}
    
    \vspace{-2mm}
\end{table}

Based on our observation and analysis of the above \bugs, we suggest two improvements to enhance the fusion methods. First, radar predictions should be integrated rather than dominated by camera predictions (Cases 1-2). Second, vehicles intending to cut in should be tracked and considered (Case 3). \mathworkoriginal already addresses the first aspect and thus has less \bugs found. Regarding the second one, for each tracked object by radar, we store their latitudinal positions at each time step. At next time step, if a vehicle's relative latitudinal position gets closer to the ego car, it will be included in the candidate pool for the leading vehicle rather than discarded. We call this new fusion method \mathworkmoving.

\begin{figure}[ht]
\centering
    \subfloat[
    $time_0$
    \label{fig:fix_1}
    ]
    {\includegraphics[width=0.11\textwidth]{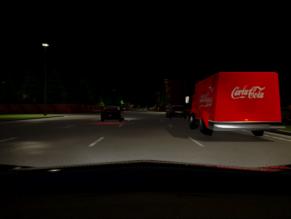}}
    \vspace{0.1mm}
    \subfloat[
    $time_1$
    \label{fig:fix_2}
    ]
    {\includegraphics[width=0.11\textwidth]{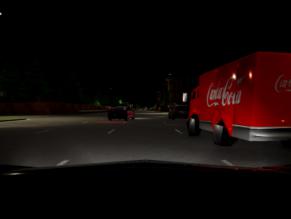}}
    \vspace{0.1mm}
    \subfloat[
    $time_2$
    \label{fig:fix_3}
    ]
    {\includegraphics[width=0.11\textwidth]{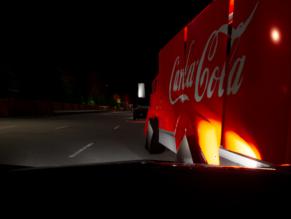}}
    \vspace{0.1mm}
    \subfloat[
    $time_2'$
    \label{fig:fix_3_fix}
    ]
    {\includegraphics[width=0.11\textwidth]{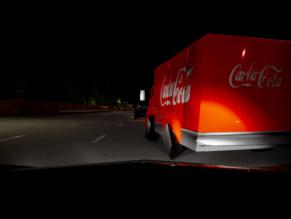}}
\caption{\small{An \bug avoided by \mathworkmoving.}}
\label{fig:fix_demo}
\vspace{-3mm}
\end{figure}

We evaluate \mathworkmoving via replacing the original fusion method with it during the pre-crash window on the previously found \bugs. As shown in \Cref{tab:bug_num_results}, at least 50\% of found \bugs can be avoided.
\Cref{fig:fix_demo} shows a \bug found on \mathworkoriginal but avoided by \mathworkmoving. At $time_0$ (\Cref{fig:fix_1}), the ego car and a red truck drive on different lanes. At $time_1$, \mathworkoriginal does not consider the truck since it just starts to invade into the current lane (the truck's radar lead is discarded at \textcircled{3} in \Cref{fig:fusion_logic_mathworkoriginal}). When the truck fully drives into the current lane, it is too late for the ego car to avoid the collision at $time_2$ (\Cref{fig:fix_3}). If \mathworkmoving is used since \Cref{fig:fix_1}, the truck would be considered a leading vehicle at \Cref{fig:fix_2} and the collision would be avoided at $time_2'$ (\Cref{fig:fix_3_fix}).
These results demonstrate the improvement of \mathworkmoving. Further, it implies that with a good  fusion method, many \bugs can be avoided without modifying the sensors or the processing units.

\RS{3}{Based on the observations of the found \bugs, we adjust the fusion method we study and enable it to avoid more than 50\% of the initial \bugs.}

%% file: body/8_discussion.tex
\section{Related Work}
\label{sec:related_work}

\noindent\textbf{\newedit{Search Based Software Engineering(SBSE).}}
\newedit{SBSE formulates a software engineering problem into a search problem and applies search-based metaheuristic optimization techniques \cite{HARMAN2001833, Harman2012, sbst11}. In our context, we apply evolutionary algorithm to search for test cases (scenarios) which can cause ego car's \bugs.}

\noindent\textbf{Scenario-Based Testing\newedit{(SBT)}.} 
In order to identify the errors of a driving automation system, comprehensive tests are being conducted by autonomous driving companies. 
The public road testing approach is the closest to a system's use case, but it is incredibly costly. It has been shown that more than 11 billion miles are required to have a 95\% confidence that a system is 20\% safer than an average human driver \cite{nummilesneeded}. Most of these miles, however, usually do not pose threats to the system under test and are thus not efficient, if not wasted.
To focus on \newedit{challenging}\sout{interesting} test cases, SBT techniques have been developed where a system is tested in \sout{some }difficult scenarios \sout{either }designed by experts or \newedit{found}\sout{searched} by algorithms. Besides, since many dangerous cases (e.g., a pedestrian crossing a street close to the ego car) cannot be tested in the real world, such tests are usually conducted in a high-fidelity simulator \cite{DBLP:journals/corr/abs-2112-00964}.
\sout{Existing works usually treat the system under test as a black-box and search for hard \specificss to trigger ego car's potential misbehaviors \cite{NEURIPS2018_653c579e, wheelerimportance2019, Abeysirigoonawardena2019Generating, ding2020learning, Kuutti2020, chenbaiming2020, korenadaptive2018, multimodaltest, testing_vision18, nsga2nn, avfuzzer, covavfuzz, autofuzz21, felbinger19, kluck19, pathgeneration, DBLP:journals/corr/abs-2112-00964}.}
\newedit{Existing works usually treat the system under test as a black-box and search for hard \specificss to trigger ego car's potential failure. Search methods leveraging evolutionary algorithms \cite{testing_vision18, nsga2nn, avfuzzer, autofuzz21}, reinforcement learning \cite{chenbaiming2020}, bayesian optimization \cite{Abeysirigoonawardena2019Generating}, and topic modeling\cite{ding2020learning} have been used.}

The \newedit{failures}\sout{errors} found can have different causes like the failure of the sensors, \sout{the issue of} the planning module, or \newedit{the modules'}\sout{some} interactions\sout{among the modules}. However, most existing works either ignore the root cause analysis or merely analyze causes for general \newedit{failures}\sout{errors}. In contrast, we focus on revealing \newedit{failures}\sout{errors} causally induced by the fusion component. 
\sout{Similar to the current work, }Abdessalem et al. \cite{interactiontest} study the \sout{interaction }failure of \newedit{an \adas's integration component which integrates the decisions of different functionalities (e.g., AEB and ACC)}\sout{several components in an \adas}. \sout{However}\newedit{In contrast}, we focus on the fusion component which integrates the data from multiple \newedit{sensors.}\sout{sensor processing modules while they study the integration module which integrates the decisions of different functionalities (e.g., AEB and ACC).} Besides, the fusion component can either be rule-based (e.g., \original) or algorithm-based (e.g., \mathwork) while the integration component studied in \cite{interactiontest} is only rule-based.

\noindent\textbf{Adversarial Attacks on Fusion.} Some recent works study how to attack the fusion component of an automated system and thus fails the system \cite{Cao2021InvisibleFB, tumultisensor2021, fusionattack}. In particular, Cao et al. \cite{Cao2021InvisibleFB} and Tu et al. \cite{tumultisensor2021} study how to construct adversarial objects that can fool both camera and Lidar at the same time and thus lead the \msf to fail. Shen et al. \cite{fusionattack} study how to send spoofing GPS signals to confuse a \msf on GPS and Lidar. In contrast to creating artificial adversarial objects or sending adversarial signals, we focus on finding \specificss under which the \newedit{faults}\sout{failure} of \msf lead to critical accidents without the presence of any malicious attacker.

\section{Threats to validity}
There remains a gap between testing in real-world and testing in a simulator. However, road testing is overly expensive and not flexible. Besides, a simulation environment allows us to run counterfactual simulations easily and attribute an \newedit{failure}\sout{error} to the fusion method used. Consequently, we focus on testing in the simulation environment.

\newnewedit{The causal graph (\Cref{fig:causal_graph}) constructed may not capture all the influential variables. To mitigate this threat, in RQ1, we run sanity check of the causal graph by checking the reproducibility of the simulation results when using the same endogenous variables.}

\newnewedit{Since we change the communication within \op and that between \op and \carla to be synchronous and deterministic (see 
Appendix C\extend{ in the extended version \cite{fusedarxiv22}}), the behavior of \op can be different from the original \op. However, this change should not influence the found \bugs since \op should perform better when it receives the latest sensor data rather than the delayed ones.}

As in \sout{previous works} \cite{paracosm, testing_vision18}, we evaluate the proposed method using the number of found scenarios leading to \bugs. However, we have observed that this metric might \newedit{double-count}\sout{double counts} similar \bugs. To mitigate this threat, we additionally use another counting metric based on the ego car’s trajectory (\Cref{sec:error_counting}). \sout{A more fine-grained metric can be developed by considering the inputs and outputs of the fusion component. We leave its exploration for future work.}

Besides, similar to previous works\cite{testing_vision18, avfuzzer}, we evaluate the studied \adas \newedit{in}\sout{on} two driving environments. Since \adas only performs the task of lane following, the complexity of its applicable 
environments is limited \newedit{and thus mostly covered in the two environments}. \sout{The two environments we consider has covered most of its applicable road situations.}

Another threat is that the hyper-parameters \sout{(e.g., the coefficients of the search objective)} are not \sout{fully} fine-tuned. However, even with the current parameters, the proposed method already outperforms the baselines. We believe that the performance of the proposed method can be \sout{further} improved by fine-tuning\sout{the hyper-parameters}. 

Furthermore, we only test \mathworkmoving on limited detected \bugs on \op. There might be corner cases that are not covered. Since we focus on \bugs finding rather than fixing, we leave a comprehensive study for future work.

Finally, the current fusion objective only applies to HLF and is only tested on two popular fusion methods in \op. Conceptually, the proposed method can generalize to the fusion component in \ads like Apollo\cite{apollo} and Autoware\cite{autoware} which use HLF components. We plan to study other types of fusion methods like MLF and LLF, as well as \msf in \ads in future work.



%% file: body/9_conclusion.tex
\section{Conclusion}
In this work, we formally define, expose, and analyze the root causes of \bugs on two widely used \msf methods in a commercial ADAS. To the best of our knowledge, our work is the first study on finding and analyzing \newedit{failures}\sout{errors} causally induced by \msf in an end-to-end system. We propose a grey-box fuzzing framework, \tool, that effectively detects \bugs. Lastly, based on the analysis of the found \bugs, we provide several learned suggestions on how to improve the studied fusion methods.

%% file: body/10_acknowledgement.tex
\section*{acknowledgement}
The majority of work was done during the internship of Ziyuan Zhong at Baidu Security X-Lab. The work is also supported in part by NSF CCF-1845893, CCF-2107405, and IIS-2221943. We also want to thank colleagues from Baidu Security X-Lab and Chengzhi Mao from Columbia University for valuable discussions.

%% file: body/appendices.tex
\appendix
\section{Background: Causality Analysis}
\label{sec:background_causality_analysis}
Over the recent years, causality analysis has gained popularity on interpreting machine learning models\cite{nncausal18, nncausal19}. Compared with traditional methods, causal approaches identify causes and effects of a model’s components and thus facilitates reasoning over its decisions. In the current work, we apply causality analysis to justify the defined \bugs are indeed \adas errors caused by the fusion method.

In causality analysis, the world is described by variables and their values. Some variables may have a causal influence on others. The influence is modeled by a set of \emph{structural equations}. The variables are split into two sets: \emph{the exogenous variables}, whose values are determined by factors outside the model, and \emph{the endogenous variables}, whose values are ultimately determined by the exogenous variables. In our context, exogenous variables can be a user specified test design for an \adas (e.g., testing the \adas for a left turn at a signalized intersection scenario), the endogenous variables can consist of the configuration for the \adas under test (e.g., the ego car's target speed), the concrete scenario (e.g., the number and locations of the NPC vehicles), and the final test results (e.g., if a collision happen).

Formally, a causal model $M$ is a pair $(S,F)$. 
$S$ is called "signature" and is a triplet $(U,V,R)$. Among them, $U$ is a set of exogenous variables, $V$ is a set of endogenous variables, and $R$ associates with every variable $Y\in U\cup V$ a nonempty set $R(Y)$ of possible values for $Y$. 
$F$ defines a set of modifiable structural equations relating the values of the variables. 

Given a signature $S=(U,V,R)$, a primitive event is a formula of the form $X=x$, for $X\in V$ and $x\in R(X)$. 
A causal formula (over $S$) is one of the form 
$[\overrightarrow{Y}\leftarrow\overrightarrow{y}]\phi$, where $\overrightarrow{Y}$ is a subset of variables in $V$ and $\phi$ is a boolean combination of primitive events. This says $\phi$ holds if $\overrightarrow{Y}$ were set to $\overrightarrow{y}$. 
We further denote $(M,\overrightarrow{u})\models\phi$ if the causal formula $\phi$ is true in causal model $M$ given context $U=\overrightarrow{u}$. $(M,\overrightarrow{u})\models(X=x)$ if the variable $X$ has value $x$ in the unique solution to the equations in $M$ given context $\overrightarrow{u}$.

We next provide the definition of causality \newedit{\cite{causalitydef15}}.
\begin{defi}
$\overrightarrow{X}=\overrightarrow{x}$ is an actual cause of $\phi$ in $(M,\overrightarrow{u})$ if the following three conditions hold: \\
AC1. $(M,\overrightarrow{u})\models(\overrightarrow{X}=\overrightarrow{x})$ and $(M,\overrightarrow{u})\models\phi$ \\
AC2. There is a partition of $V$ into two disjoint subsets $\overrightarrow{Z}$ and $\overrightarrow{W}$ (so that
$\overrightarrow{Z}\cap \overrightarrow{W}=\emptyset$) with $\overrightarrow{X} \subseteq \overrightarrow{Z}$ and a setting $\overrightarrow{x}'$ of the variables in $\overrightarrow{X}$ such that if $(M,\overrightarrow{u})\models(\overrightarrow{W}=\overrightarrow{w})$ then 
\[
(M,\overrightarrow{u})\models[\overrightarrow{X}\leftarrow \overrightarrow{x'}, \overrightarrow{W}\leftarrow \overrightarrow{w}]\neg\phi
\]
AC3. $\overrightarrow{X}$ is minimal; no subset of $\overrightarrow{X}$ satisfies AC1 and AC2.
\end{defi}

AC1 means both $\overrightarrow{X}=\overrightarrow{x}$ and $\phi$ happen. 
AC3 is a minimality condition. It makes sure only elements within $\overrightarrow{X}$ that are essential are considered as part of a cause. 
Without AC3, if studying hard causes good grade then so is studying hard and being tall.
AC2 essentially says when keeping everything else (except $\overrightarrow{X}$ and those influenced by it) exactly the same, there is a $\overrightarrow{x}'$ such that it can make $\phi$ does not hold.

In the current work, given an error happens (i.e., $\phi$ holds), we want to determine if the fusion method ($\overrightarrow{X}=\overrightarrow{x}$) is the actual cause. To achieve this, we want to identify scenarios such that AC1-AC3 are all satisfied. In fact, our defined \bugs satisfy them.

Since $\overrightarrow{X}$ only consists of one variable, the minimality condition AC3 is satisfied automatically. Given a collision scenario (i.e., $\phi$ is true), AC1 is also trivially satisfied with the fusion method $\overrightarrow{x}$ used during the collision. The issue remains on how to achieve AC2, which is the main condition we analyze in \Cref{sec:root_cause_analysis}.

\section{Radar Implementation for \op in \carla}
The official \op bridge implementation does not implement radar in the simulation environment. To evaluate the fusion component of \op, we create a radar sensor in \carla following Delphi Electronically Scanning RADAR \cite{radarspec} ($174$m range and $30$ horizontal degrees) and send processed radar data at 20Hz. Since the radar interface in CAN of \op takes in pre-processed $16$ tracks including relative x, relative y, and relative speed, the raw radar information in \carla is pre-processed using DBSCAN \cite{dbscan} (with eps$=0.5$ and min samples$=5$), a widely used for raw radar data pre-processing \cite{radarsurvey}.

\section{Settings for \carla and Changes made on \op for Improved Simulation Determinism and Reproducibility}
\label{sec:adaption_of_openpilot}
By default, the communications among \carla client (\op in our case), traffic manager (controlling NPC vehicles), and \carla server (controlling the world environment) use asynchronous communication. 
The \carla world also has a non-stationary "time step" between every two updates. In particular, the virtual time between two updates of the world is set to be equal to the real-world computing time. 
Besides, \op is a real-time system and its internal communications among its modules are asynchronous. 
These default designs introduce non-determinism to a simulation's result which becomes dependent on the underlying system's state. 
This also makes reproducing an accident of \op in \carla very difficult. 

To conquer these difficulties, we apply the following configuration settings and changes. Note that for simplicity, in the current work, we collectively call the changes we have made as setting the \carla communication configurations and the \op communication configurations to be deterministic and synchronous.

First, we adopt synchronous mode for \carla and \tm, set a random seed to 0 for \tm, and use a fixed virtual time step of 0.01 seconds. These allow the scenarios created to be deterministic. Second, we adopt some design choices of Testpilot\cite{blackbox19} which was developed based on \op 0.5 (a very early version) and achieved complete synchronization between \carla and \op. 
In particular, we modify \op to use the simulator's virtual time passed from the bridge between the simulator and the controller rather than using the real world time.
Besides, we modify the communication among the modules inside \op to be synchronous. 
Note that if \op uses real time, \carla will be not able to catch up with the speed of \op and thus lead to significant communication delay. What's more, before each scenario starts, we allow \op to take in sensor \newedit{outputs}\sout{input} from \carla for 4 virtual seconds to allow all of its modules up and running. 
This step is necessary since on different machines, it takes different virtual time for \op to start all the modules.
With these changes, the simulation is more deterministic, and \bugs can be easily reproduced across runs and machines. 


\section{Adaption of Mathwork Radar-Camera Fusion}
\label{sec:adaption_of_mathwork}
The original Mathwork implementation is in Matlab, we reimplement it in Python. Besides, the Mathwork implementation uses an Extended Kalman Filter since the radar measurements are non-linear. Since \op takes a transformed linear radar measurements (i.e. the relative position is in the Cartersion coordinate rather than angular coordinate), we use a Linear Kalman Filter instead. Besides, the speed of $y$ axis (in terms of the ego car's coordinate) is not supported by the \op interface so  we leave this field as a constant in the Kalman Filter modeling. These changes can potentially introduce \bugs that may not appear in the original implementation.




\section{Driving Environment Details}
\label{sec:search_space_details}
\newedit{
We utilize two \lss named \Sa and \Sb in our study. \Sa is a straight local road and \Sb is a left curved highway road. An illustration is shown in \Cref{fig:scenarios_illustration} (not all NPC vehicles are shown).
In each driving environment, the search space has $76$ dimensions since it has six NPC vehicles($6*11=66$ searchable fields), two searchable lighting fields, and eight searchable weather fields. 
Lighting condition is controlled by sun azimuth angle and sun altitude angle. Weather consists of the following fields: cloudiness, precipitation, precipitation deposits, wind intensity, fog density, fog distance, wetness, and fog falloff, the detailed explanation of each field can be found on \carla Python API \cite{carla}. Each NPC vehicle has a model type field and five waypoints, each of which consists of two fields (speed and changing lane). Thus each NPC vehicle has $11$ ($=1+5\times 2$) dimensions to search for. 
}

\begin{table}[ht]
\scriptsize
    \centering
    
    \begin{tabular}{l|l|l|l|l}
    \toprule
        object & sub-object & property & data type & range\\
    \toprule
        $\textrm{vehicle}_i$ & model type & index & discrete & \{0,...,27\} \\
    \cmidrule{2-5}
                  i=1,...,6 & $\textrm{waypoint}_{ij}$ & speed & continuous & [-100,50] \\
                  & j=1,...,5           & lane change & discrete & \{0,1,2\} \\
    \midrule
        background & lighting & sun azimuth angle & continuous & [0,360]\\  
                 & & sun altitude angle & continuous & [-90,90] \\ 
     \cmidrule{2-5}
         & weather & cloudiness & continuous & [0,100] \\ 
                & & precipitation & continuous & [0,80] \\
                & & precipitation deposits & continuous & [0,80] \\ 
                & & wind intensity & continuous & [0,50] \\
                & & fog density & continuous & [0,15] \\
                & & fog distance & continuous & [0,100] \\
                & & wetness & continuous & [0,40] \\
                & & fog falloff & continuous & [0,2] \\
    \bottomrule
    \end{tabular}
    \caption{\small{\newedit{Details of the search space.}} \label{tab:search_fields}}
\end{table}

\newedit{We next provide the details for the search range for each field. The two driving environments S1 and S2 have the same fields in their search space. \Cref{tab:search_fields} shows all the fields in the search space. There are six vehicles (indexed i=1,...,6, respectively). For each vehicle, it has a model type field as well as five waypoint fields (indexed j=1,...,5). There are 27 different models to sample from (23 for S2 since it does not have cyclists and motorcycles). Note that each waypoint field means a behavior change. The five waypoints are evenly distributed in the specified virtual simulation time (in our case, the time is 20 seconds). Each waypoint has two fields: speed and lane change. Speed means what new speed the vehicle should change to and keep for the next time interval. The speed can be set to as low as 0 (when -100 is used) and as high as 50\% more than the current road's speed limit (when 50 is used). The maximum allowed speed of the auto-driving car is set to $45$ miles/hr \newedit{($\approx 20.1$m/s)} on the highway road (\Sb) and $35$ miles/hr \newedit{($\approx 15.6$m/s)} on the local road (\Sa). For lane change, 0 represents no change, 1 represents changing to the left lane if available, and 2 represents changing to the right lane if available. The ranges for the lighting fields and the weather fields are designed based on the ranges in CARLA. The details can be found in the CARLA's official documentation \cite{carla}.}

\section{Discussion on driving environments and the practical capability of \op}
\label{sec:op_capability}
\newedit{
The studied driving environments are within the capabilities of \op. We clarify this point in the following three aspects:}

\noindent\newedit{\textbf{Safety Guard:} \op has a safety guard module called 'Forward Collision Warning'(FCW). We have FCW enabled throughout our experiment. However, all the \bugs we found didn't trigger such FCW, which means \op's safety guard does not detect such hazards. Moreover, due to the sudden hazard, the drivers may have very limited time to take over the driving properly.}

\noindent\newedit{\textbf{Detection Capability:} \op indeed has detected the lead vehicles by its own detection model, though it is different from traditional object detection which tells the size, class, and the location of the obstacles. Our findings root from the finer-grained analysis of \op’s detection results.}

\noindent\newedit{\textbf{No Lane Change for \op:} Throughout our experiment, we make the ego car not change lanes so it keeps on following the current lane. Thus, the \bugs we found do not involve any lane-changing situations of the ego car.}

\newedit{To summarize, we carefully design the driving environments to make sure they comply with the practical capabilities of \op. Despite such restrictions, we still identified dangerous situations that were worth certain attention.
}

\section{Sensitivity Analysis of the pre-crash period's duration}
\label{sec:precrash_ablation}

\begin{table}[ht]
    \scriptsize
    \centering
    \caption{\small{mean and standard deviation of \# distinct \bugs.}\label{tab:ablation_pre_crash}}
    \begin{tabular}{l|l|l}
    \toprule
        Algorithms & m=$1.5$s & m=$3.5$s \\
        
    \toprule
      \alg & $48.6\pm 1.5$ & $58.3\pm 3.1$ \\
    \midrule
      \GA & $32.7\pm 2.9$ & $43.0\pm 2.6$ \\
    \bottomrule
    \end{tabular}
    
\end{table}

\newedit{We next explore the sensitivity of the proposed method under different pre-crash period $m$. In particular, we compared \alg and \GA when setting $m$ to $1.5$s and $3.5$s (We set $m$ to $2.5$s for the results in our main text). As shown in \Cref{tab:ablation_pre_crash}, under all the settings, on average of three runs, \alg has found more distinct \bugs than \GA. The result shows the \alg is superior under different $m$. Another observation is that as $m$ increases, the numbers of \bugs and distinct \bugs increase. This is because with larger $m$, the fusion method will be replaced by the best sensor fusion method earlier and thus \op is more likely to avoid the original collision. Consequently, more \bugs and distinct \bugs will be reported. }